\documentclass[10pt,journal,compsoc]{IEEEtran}
\ifCLASSOPTIONcompsoc
  \usepackage[nocompress]{cite}
\else
  \usepackage{cite}
\fi

\ifCLASSINFOpdf
 \else
 \fi
\usepackage{amsmath}
\usepackage{tikz}
\usepackage{array}
\usepackage{times}
\usepackage{subfig}
\usepackage{helvet}
\usepackage{subfig}
\usepackage{amssymb}
\usepackage{amsmath}
\usepackage{amssymb}
\usepackage{courier}
\usepackage{multirow}
\usepackage{mathrsfs}
\usepackage{graphicx}
\usepackage{enumitem}
\usepackage{graphicx}
\usepackage{blindtext}
\usepackage{algorithm}
\usepackage{algorithmic}
\usepackage{lineno}
\usepackage[marginal]{footmisc}
\usepackage{booktabs}
\usepackage{endnotes}
\usepackage{amsthm}
\theoremstyle{plain}
\theoremstyle{definition}
\renewcommand{\paragraph}[1]{\vspace{2ex}\noindent\textbf{#1}.}
\usepackage[utf8]{inputenc}
\usepackage[english]{babel}
\usepackage{epstopdf}
\usepackage{array}
\usepackage{multirow}
\usepackage{booktabs}

\newtheorem{definition}{Definition}

\usepackage{bbm}
\usepackage{colortbl}  
\usepackage{xcolor}
\usepackage{array} 

\def\BibTeX{{\rm B\kern-.05em{\sc i\kern-.025em b}\kern-.08em
    T\kern-.1667em\lower.7ex\hbox{E}\kern-.125emX}}

\ifCLASSOPTIONcompsoc
  \usepackage[nocompress]{cite}
\else
  \usepackage{cite}
\fi

\ifCLASSINFOpdf
\else
\fi

\usepackage{tikz}
\usepackage{mathrsfs}
\usepackage{amssymb}
\usepackage{bm}
\usepackage{amsmath}
\usepackage{multirow}
\usepackage{comment}
\usepackage{ragged2e}
\begin{document}
\title{A Survey of Knowledge Graph Reasoning on Graph Types: Static, Dynamic, and Multi-Modal}
\author{Ke Liang,
        Lingyuan Meng,
        Meng Liu,
        Yue Liu,
        Wenxuan Tu, Siwei Wang, Sihang Zhou,\\
        Xinwang Liu$^{\dag}$, ~\IEEEmembership{Senior~Member,~IEEE},
        Fuchun Sun, ~\IEEEmembership{Fellow,~IEEE}
\IEEEcompsocitemizethanks{
\IEEEcompsocthanksitem $^{\dag}$ Corresponding Author.
\IEEEcompsocthanksitem Ke Liang, Lingyuan Meng, Meng Liu, Yue Liu, Wenxuan Tu, Siwei Wang, and Xinwang Liu are with the School of Computer, National University of Defense Technology, Changsha, 410073, China. E-mail: {xinwangliu@nudt.edu.cn}.
\IEEEcompsocthanksitem Sihang Zhou is with the College of Intelligence Science and Technology, National University of Defense Technology, Changsha, 410073, China.
\IEEEcompsocthanksitem Fuchun Sun is with the Department of Computer Science and Technology, Tsinghua University, Beijing, 100084, China.}
\thanks{This work has been submitted to the IEEE for possible publication. Copyright may be transferred without notice, after which this version may no longer be accessible.}}

%
%

\markboth{}
{Shell \MakeLowercase{\textit{et al.}}: Bare Advanced Demo of IEEEtran.cls for IEEE Computer Society Journals}

\IEEEtitleabstractindextext{%
\begin{abstract}
\justifying
Knowledge graph reasoning (KGR), aiming to deduce new facts from existing facts based on mined logic rules underlying knowledge graphs (KGs), has become a fast-growing research direction. It has been proven to significantly benefit the usage of KGs in many AI applications, such as question answering, recommendation systems, and etc. According to the graph types, existing KGR models can be roughly divided into three categories, \textit{i.e.,} static models, temporal models, and multi-modal models. Early works in this domain mainly focus on static KGR, and recent works try to leverage the temporal and multi-modal information, which are more practical and closer to real-world. However, no survey papers and open-source repositories comprehensively summarize and discuss models in this important direction. To fill the gap, we conduct a first survey for knowledge graph reasoning tracing from static to temporal and then to multi-modal KGs. Concretely, the models are reviewed based on bi-level taxonomy, \textit{i.e.,} top-level (graph types) and base-level (techniques and scenarios). Besides, the performances, as well as datasets, are summarized and presented. Moreover, we point out the challenges and potential opportunities to enlighten the readers. The corresponding open-source repository is shared on GitHub {https://github.com/LIANGKE23/Awesome-Knowledge-Graph-Reasoning}.
\end{abstract}

\begin{IEEEkeywords}
Knowledge Graph Reasoning, Knowledge Graph, Temporal Knowledge Graph, Multi-Modal Knowledge Graph.
\end{IEEEkeywords}}

\maketitle
\IEEEdisplaynontitleabstractindextext
\IEEEpeerreviewmaketitle

\ifCLASSOPTIONcompsoc
\IEEEraisesectionheading{\section{Introduction}\label{sec:introduction}}
\else
\section{Introduction}
\label{sec:introduction}
\fi

\begin{figure}[!t]
\centering
\includegraphics[width=0.48\textwidth]{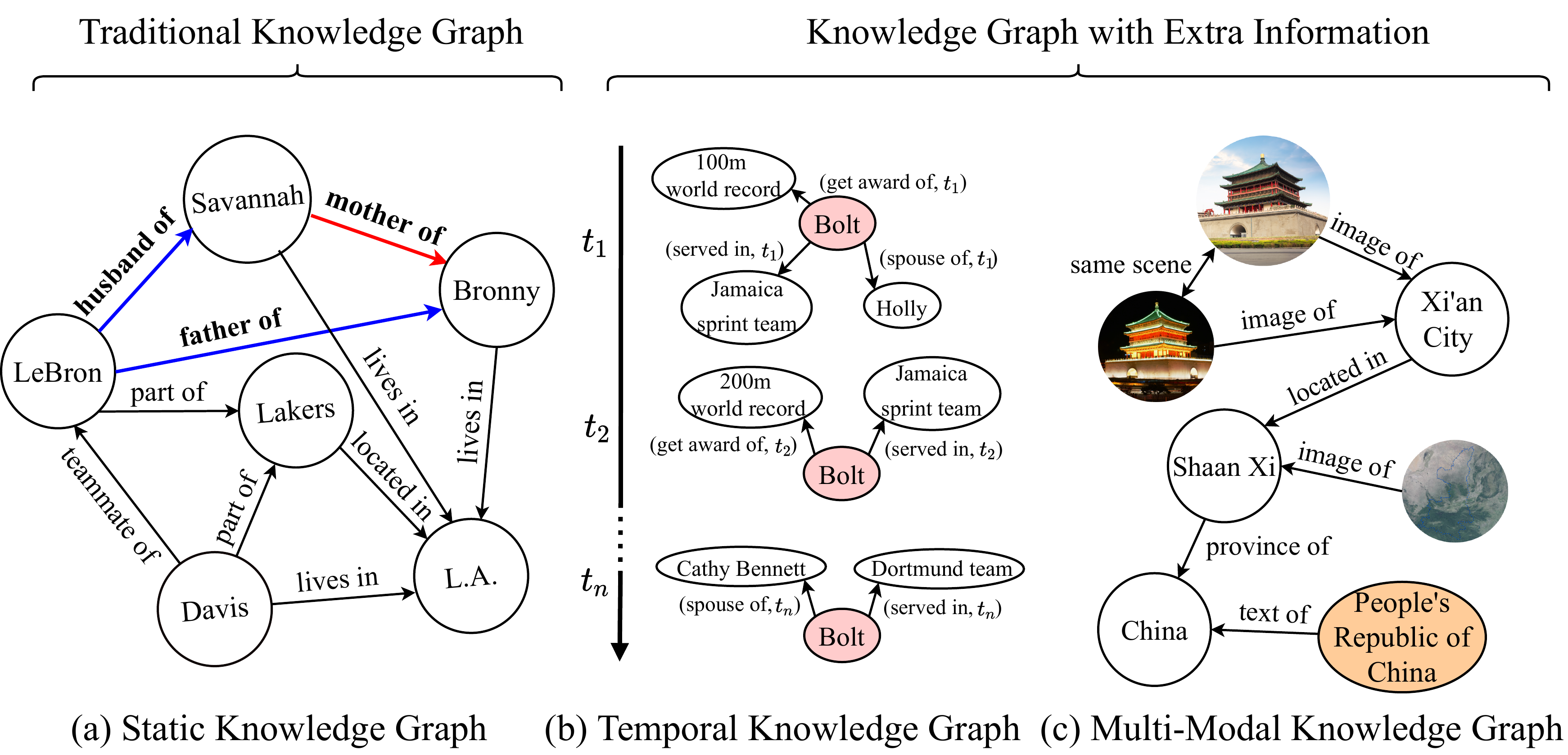} 
\caption{Examples of three categories of the knowledge graphs, \textit{i.e.,} static, temporal, and multi-modal knowledge graph.}
\vspace{-0.4 cm}
\label{comp}
\end{figure}

\IEEEPARstart{H}{umans} learn skills from two main sources, \textit{i.e.,} specialized books and working experiences. For example, a good doctor needs to get knowledge from school and practice experiences from the hospital. However, most existing artificial intelligence (AI) models only imitate the learning procedure from experiences while ignoring the former \cite{PAMI1, PAMI2, PAMI3, PAMI4}, thus making them less explainable and worse performances. Knowledge graphs (KGs) \cite{KGPAMI}, which store the human knowledge facts in intuitive graph structures \cite{KGESymCL}, are treated as potential solutions these years. While, the construction of KGs is a dynamic and continuous procedure, thus most KGs suffer from incomplete issues, hindering their effectiveness in KG-assisted applications, such as question answering \cite{KGESURVEY}, recommendation system \cite{wong2021improving}, etc. To alleviate the problem, knowledge graph reasoning (KGR) has drawn increasing attention these years. It aims to infer missing facts from existing ones in KGs. Taking Figure \ref{comp} (a) as the target KG, KGR models are expected to derive the logic rules \emph{(A, father of, B)}$\wedge$\emph{(A, husband of, C)}$\rightarrow$\emph{(C, mother of, B)}, and then further infer the missing fact \emph{(Savannah, mother of, Bronny)}. 

According to the information types in KGs, current KGs can be roughly divided into three categories, \textit{i.e.,} static KGs, temporal KGs, and multi-modal KGs, as shown in Figure \ref{comp}. Traditional KGs only contain static uni-modal facts, which is simple but effective for developing general basic KGR models. However, they still cannot fully describe real-world scenarios, which consist of information from various sources. Thus, recent KGs (\textit{i.e.,} temporal KGs and multi-modal KGs) are constructed by integrating extra temporal and multi-modal information based on static KGs, which is more practical and closer to the real world. However, no matter which types of KGs they are, the incomplete problem still exists. Therefore, various advanced KGR models have been continuously developed and studied these years for better reasoning performance. Of course, it is worth noting that the core issues of KGR models for different KG types are different. Specifically, static KGR models focus on the general representation learning capacity. While, how to fuse extra information well is the key to the temporal and multi-modal KGR models. Moreover, for a more comprehensive and systematical review, two sub-taxonomies, \textit{i.e.,} reasoning techniques and reasoning scenarios, are further discussed within each graph type. 

\begin{figure*}[t]
\centering
\includegraphics[width=\textwidth]{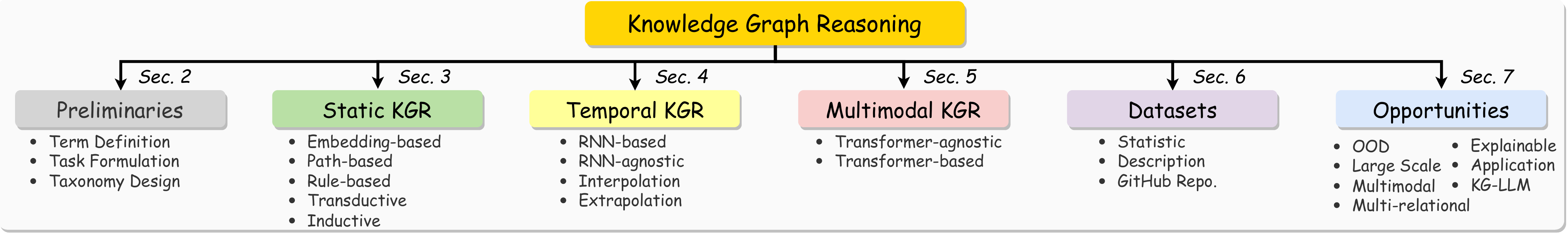}
\vspace{-0.6 cm}
\caption{Overview framework of the survey.}
\label{OverFrame}  
\vspace{-0.5 cm}
\end{figure*}

\begin{table}[t]
\caption{Comparison between different KGR surveys.}
\vspace{-0.2 cm}
\fontsize{13}{16}\selectfont 
 \label{comps}
\resizebox{\linewidth}{!}{
\begin{tabular}{ccccccccc}
\hline
\textbf{Survey}                 & \textbf{\cite{survey1}} & \textbf{\cite{survey2}} & \textbf{\cite{survey3}} & \textbf{\cite{chen2022overview}} & \textbf{\cite{chen2023generalizing}} & \textbf{\cite{temporalSurvey}} & \textbf{\cite{MMKGS}} & \textbf{Ours} \\ \hline
Static KGR             & $\checkmark$  &  $\checkmark$ & $\checkmark$  & $\checkmark$  & $\checkmark$   &   &   &  \textbf{$\checkmark$}    \\
Embedding-based Model  &  $\checkmark$ & $\checkmark$  & $\checkmark$  &  $\checkmark$ & $\checkmark$   &   &   &  $\textbf{\checkmark}$     \\
Path-based Model   & $\checkmark$  &  $\checkmark$ &  $\checkmark$  & $\checkmark$  &   &   &   &    $\textbf{\checkmark}$   \\
Rule-based Model &  $\checkmark$ & $\checkmark$  & $\checkmark$  &  $\checkmark$ &   &   &   &  $\textbf{\checkmark}$     \\
Transductive Scenario  &   &   &  &   &   &   &   &  $\textbf{\checkmark}$     \\
Inductive Scenario     &   &   &   &   &  $\checkmark$ &   &   &    $\textbf{\checkmark}$   \\\hline
Temporal KGR           &   &   &   &   &   &  $\checkmark$ &   &    $\textbf{\checkmark}$   \\
RNN-based Model  &  &  &   &   &   &  $\checkmark$ &   &  $\textbf{\checkmark}$     \\
RNN-agnostic Model   &   &   &   &   &   & $\checkmark$  &   &    $\textbf{\checkmark}$  \\ 
Interpolation Scenario &   &   &   &   &   & &   &   $\textbf{\checkmark}$    \\
Extrapolation Scenario &   &   &   &   &   &   &   &    $\textbf{\checkmark}$   \\\hline
Multi-Modal KGR         &   &   &   &   &   &   & $\checkmark$ &  $\textbf{\checkmark}$     \\ 
Non-Transformer Model  &   &   &   &  &   &   &   &  $\textbf{\checkmark}$     \\
Transformer Model   &   &   &   &   &   &   &   &    $\textbf{\checkmark}$  \\\hline
\end{tabular}
}
\vspace{-0.35 cm}
\end{table}

There are several survey papers for KGR. Most of them only focus on static KGR but omit the recent progress in other KGs, \textit{i.e.,} temporal KGs and multi-modal KGs. \cite{survey1} first categorizes KGR tasks into symbolic and statistical reasoning. Besides, \cite{survey2} divides KGR models into three types, \textit{i.e.,} symbolic, neural, and hybrid. After that, \cite{survey3} and \cite{chen2022overview} propose more fine-grained categorizations for logic-based and embedding-based KGR models. More recently, \cite{chen2023generalizing} analyzes the generalization ability of KGR models to unseen elements. As for temporal KGR, \cite{temporalSurvey} reviews existing models based on how fact timestamps capture the temporal dynamics. But it does not clearly distinguish the interpolation and extrapolation scenarios for temporal KGR. As the most influential multi-modal KG survey, \cite{MMKGS} focuses more on construction and applications than reasoning over them. Compared to those existing surveys, we conduct a more comprehensive survey for knowledge graph reasoning (See Table \ref{comps}), tracing from static to temporal and then to multi-modal KGs. More specifically, a bi-level taxonomy is leveraged for review, \textit{i.e.,} top level (graph types) and base level (techniques and scenarios). In particular, we carefully discuss reasoning scenarios for the reviewed models, \textit{i.e.,} transductive and inductive scenario for static KGR, and interpolation and extrapolation scenario for temporal KGR.

To summarize, we are the first to thoroughly survey the existing KGR models over different graph types, including traditional KGs (static KGs) and KGs with extra information (temporal KGs and multi-modal KGs). Specifically, we first introduce the preliminary (Sec. 2). Next, we systematically review the recent state-of-the-art KGR models from Sec. 3 to Sec.5 based on the bi-level taxonomy and performances. Later on, we organize and collect typical KGR datasets in Sec. 6. Then, the challenges and potential opportunities are pointed out in Sec. 7. Finally, Sec. 8 concludes the paper. To enhance the value of this survey, we summarize the main contributions as follows:
\begin{itemize}
    \item \textbf{Comprehensive Review}. We comprehensively investigate typical KGR models based on a bi-level taxonomy, \textit{i.e.,} top-level (graph types), and base-level (techniques, scenarios). Three graph types (\textit{i.e.,} static, temporal, multi-modal KGs), fourteen techniques, and four reasoning scenarios are included, which provides systematical reviews for KGR.
  
     \item \textbf{Insightful Analysis}. We analyze the strengths and weaknesses of the existing KGR models and their suitable scope, which will provide the readers with useful guidance to select the baselines for their research. 
     
    \item \textbf{Potential Opportunity}. We summarize the challenges of knowledge graph reasoning and point out some potential opportunities which will enlighten the readers.
    
    \item \textbf{Open-source Resource}. We share the collection of \textbf{180} state-of-the-art KGR models (\textit{i.e.,} papers and codes) and \textbf{67} typical datasets on GitHub \footnote{https://github.com/LIANGKE23/Awesome-Knowledge-Graph-Reasoning}.
\end{itemize}

\section{Preliminary}
In this section, we first formally define static, temporal, and multi-modal knowledge graphs. Then, the reasoning tasks over the different types of KGs and scenarios are formulated. At last, we introduce the taxonomy criterion of KGR models.
\begin{table}[t]
\centering
\small
\caption{Notation summary}
\vspace{-0.2 cm}
\resizebox{\linewidth}{!}{
\begin{tabular}{cc}
\hline
\textbf{Notation} & \textbf{Explanation} \\\hline
$\mathcal{SKG}$     &    Static knowledge graph    \\
$\mathcal{TKG}$       &   Temporal knowledge graph    \\
$\mathcal{MKG}$      &   Multi-modal knowledge graph    \\
$\mathcal{E}$ & Entity set\\
$\mathcal{R}$ & Relation set\\
$\mathcal{F}$ & The set of facts, \textit{i.e.,} edges\\
$\mathcal{T}$ & The set of the time stamps\\
$\mathcal{F}_{t}$ & The set of facts at time $t$\\
$(e_h,r,e_t)$ & Fact triplet of the head, relation, tail.\\
$(e_h, r, e_t, t)$ & Fact quadruple of the head, relation, tail, timestamp\\ 
$(e_{h}^{q}, r^{q}, e_{t}^q)$ & Queried fact triplet of head, relation, tail\\
$\textbf{e}$ & Embedding of entity\\
$\textbf{r}$ & Embedding of relation\\
$\textbf{t}$ & Embedding of timestamp\\\hline
\end{tabular}}
\vspace{-0.25 cm}
\label{NOTATION_TABLE} 
\end{table}

\subsection{Definition and Notation}
Knowledge graphs (KGs) can be viewed as graphical knowledge bases, thus inheriting most functions of traditional knowledge bases \cite{richens1956preprogramming}, such as storing, indexing, etc, but in a more intuitive manner. Existing KGs can be roughly divided into three types, \textit{i.e.,} static, temporal, multi-modal KGs. Following previous literature, the definitions of them are declared below and the notations are summarized in Table \ref{NOTATION_TABLE}.

\begin{definition}
\textbf{Static Knowledge Graph}.
\emph{Static knowledge graph (KG) is defined as $\mathcal{SKG}$ \emph{=} $\{\mathcal{E}$, $\mathcal{R}$, $\mathcal{F}\}$, where $\mathcal{E},\ \mathcal{R}$, and $\mathcal{F}$ represent the sets of entities, relations, and facts. The fact is in a triplet format $(e_h, r, e_t) \in \mathcal{F}$, where $e_h, e_t \in \mathcal{E}$, and $r \in \mathcal{R}$ between them. Note that static KGs are known as traditional KGs in \cite{survey1}. Phrase \emph{"static"} is to distinguish it from other KG types.}
\end{definition}

\begin{definition}
\textbf{Temporal Knowledge Graph}.
\emph{Temporal knowledge graph (KG) is defined as a sequence of static KGs at different timestamps $\mathcal{TKG}$\emph{=}$\{\mathcal{SKG}_1, \mathcal{SKG}_2, \mathcal{SKG}_3,\cdots, \mathcal{SKG}_t\}$. The KG snapshot at timestamp $t$ is defined as $\mathcal{SKG}_t$\emph{=}$\{\mathcal{E}$, $\mathcal{R}$, $\mathcal{F}_t\}$, where $\mathcal{E}, \mathcal{R}$ are the sets of entities and relations, $\mathcal{F}_t$ is the set of facts at timestamp $t \in \mathcal{T}$. The quadruple fact $(e_h,r,e_t,t)$ represents that relation $r$ exists between head $e_h$ and tail $e_t$ at timestamp $t$.}
\end{definition}

\begin{figure}[t]
\centering
\includegraphics[width=0.48\textwidth]{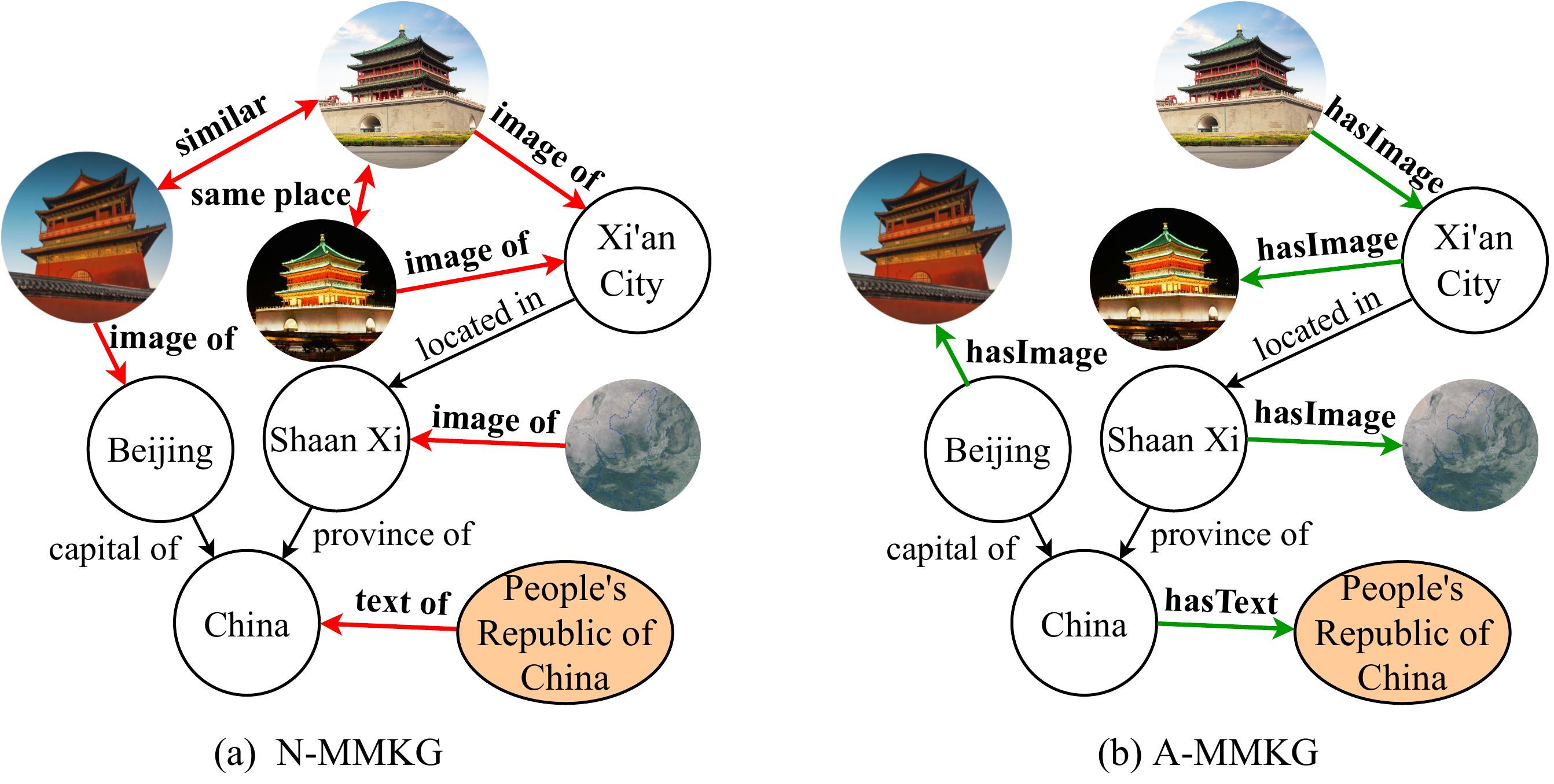} 
\vspace{-0.2 cm}
\caption{Comparison between two types of multi-modal knowledge graphs. \emph{N-MMKG} represents the multi-modal data as entities, while \emph{A-MMKG} represents multi-modal data as new attributes.}
\label{N-A-MMKG}
\vspace{-0.25 cm}
\end{figure}

\begin{definition}
\textbf{Multi-Modal Knowledge Graph}. 
\emph{Multi-modal knowledge graph (KG) $\mathcal{MKG}$ is composed of knowledge facts where more than one modalities exist. According to the representation mode of other modal data, there are two multi-modal KG \cite{survey_mmkg}, \textit{i.e.,} \emph{N-MMKG} and \emph{A-MMKG} (See Figure \ref{N-A-MMKG}).}
\end{definition}

\vspace{-0.05cm}
\subsection{Task Formulation}
Knowledge graph reasoning (KGR) aims to deduce new facts from existing facts based on the derived underlying logic rules. According to graph types, KGR can be categorized into three tasks, \textit{i.e.,} static, temporal, and multi-modal KGR.  Among them, since extra time and visual information are integrated into temporal and multi-modal KGs, respectively, there exist slight differences in task formulation compared to static KGR. Besides, two groups of terms of the reasoning scenarios are also introduced, \textit{i.e.,} transductive $\&$ inductive scenarios and interpolation $\&$ extrapolation scenarios, for a better understanding of our taxonomy.

\paragraph{Static Knowledge Graph Reasoning}
Given a static KG $\mathcal{SKG}$ \emph{=} $\{\mathcal{E}$, $\mathcal{R}$, $\mathcal{F}\}$, KGR aims to exploit the existing facts to infer queried fact $(e_h^{q}, r_{q}, e_t^{q})$ based on the likelihood calculated by scoring functions. According to the type of missing elements, there are three sub-tasks, \textit{i.e.,} head reasoning $(?, r^{q}, e_{t}^{q})$, tail inferring $(e_h^{q}, r^{q},?)$, and relation inferring $(e_{h}^{q}, ?, e_{t}^{q})$.

\paragraph{Temporal Knowledge Graph Reasoning}
Given a temporal KG $\mathcal{TKG}$ \emph{=} $\{\mathcal{SKG}_1, \mathcal{SKG}_2, \cdots, \mathcal{SKG}_t\}$, where $\mathcal{SKG}_t$ \emph{=} $\{\mathcal{E}$, $\mathcal{R}$, $\mathcal{F}_t\}$ and timestamp $t \in \mathcal{T}$, KGR aims to infer the quadruple fact $(e_h^{q}, r^{q}, e_t^{q}, t^{q})$. Similar to static KGR, there also exist three sub-tasks at specific timestamp $t^{q}$.

\paragraph{Multi-Modal Knowledge Graph Reasoning}
Multi-modal KGR is similar to the other two KGR types, \textit{i.e.,} inferring missing facts. Besides, extra fusion modules are usually required to leverage the multi-modal information for better reasoning performance.

\begin{figure}[t]
\centering
\includegraphics[width=0.48\textwidth]{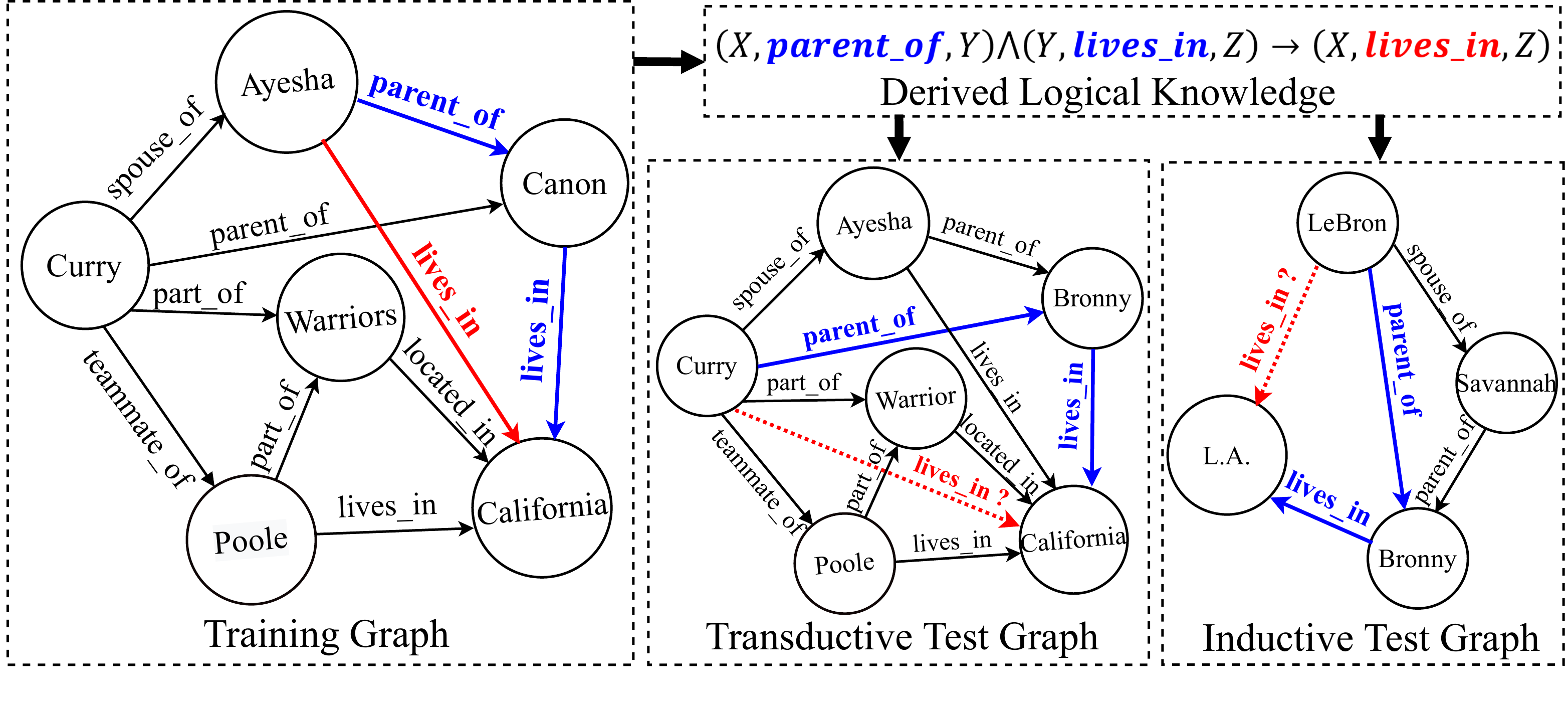}
\vspace{-0.25 cm}
\caption{Illustration of transductive and inductive reasoning. In the transductive scenario, entities in test graphs are all seen during the training procedure. While as for the inductive scenario, unseen entities may exist in test graphs.}
\label{T_IComp}  
\vspace{-0.1 cm}
\end{figure}

\paragraph{Transductive and Inductive Reasoning Scenarios}
According to the visibility of queried entities and relations during training, there are two types of reasoning scenarios (See Figure \ref{T_IComp}), \textit{i.e.,} transductive and inductive scenarios. Within transductive scenarios, entities and relationships in the queried fact are all seen in the given KG, \textit{i.e.,} $e_h^{q}, e_t^{q} \in \mathcal{E}$ and $r^{q} \in \mathcal{R}$. As for inductive scenarios, the candidates for $e_h^{q}, e_t^{q}$ and $r^{q}$ may be beyond the given KG. These two scenarios are usually discussed in static KGR.
\begin{figure}[t]
\centering
\includegraphics[width=0.48\textwidth]{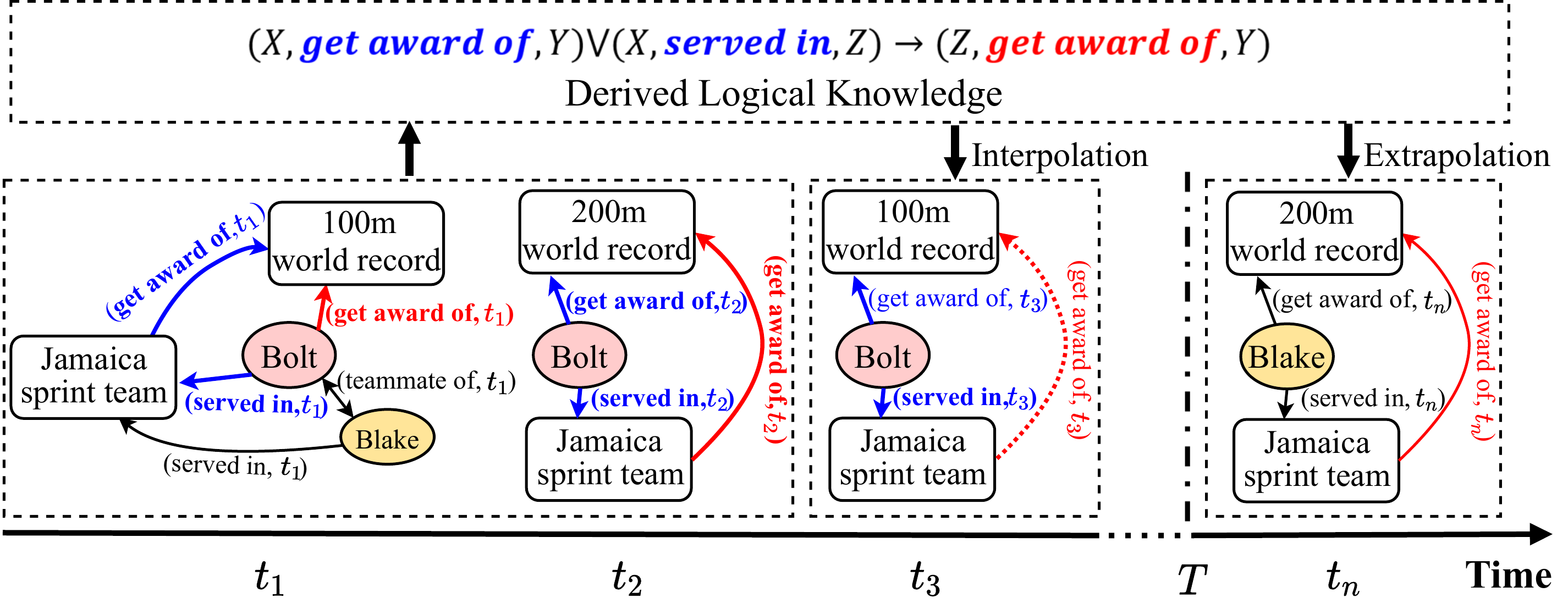}
\vspace{-0.2 cm}
\caption{Illustration of interpolation and extrapolation reasoning. The timestamp $t$ for knowledge graph reasoning in the interpolation scenario is seen in the past ($0 \leq t \leq T$). While the queried facts in the future ($t \geq T$) for the extrapolation scenario.}
\label{E_I_Comp}  
\vspace{-0.2 cm}
\end{figure}

\paragraph{Interpolation and Extrapolation Reasoning Scenarios}
According to the occurrence time of the queried fact $t^q$, temporal KGR can be divided into two categories (See Figure \ref{E_I_Comp}), \textit{i.e.,} interpolation and extrapolation scenarios. Concretely, given a temporal KG with the timestamps ranging from time $0$ to time $T$, the interpolation reasoning aims to infer the queried facts for time $t$, where $0 \leq t \leq T$; besides, the extrapolation reasoning aims to infer the queried facts for time $t$, where $t \geq T$. These two scenarios are usually discussed in temporal KGR.
\begin{figure}[b]
\centering
\vspace{-0.1 cm}
\includegraphics[width=0.42\textwidth]{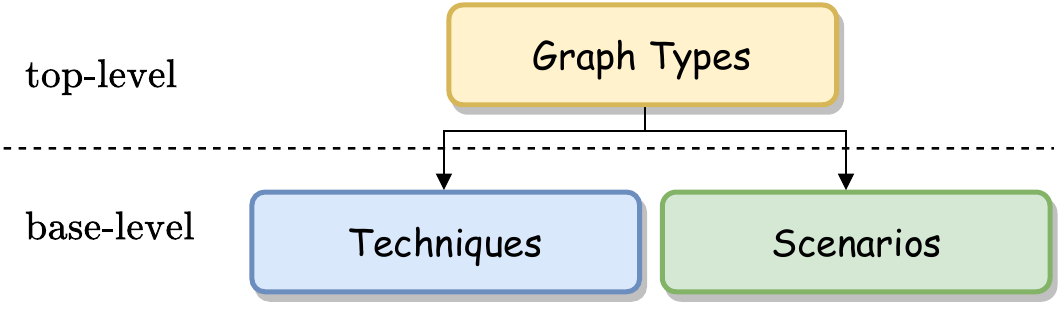}
\vspace{-0.1 cm}
\caption{The bi-level taxonomy of reviewed KGR models.}
\vspace{-0.2 cm}
\label{tax}  
\end{figure}

\subsection{Taxonomy Design}
\begin{figure}[t]
\centering
\setlength{\belowcaptionskip}{-5pt}
\includegraphics[width=0.48\textwidth]{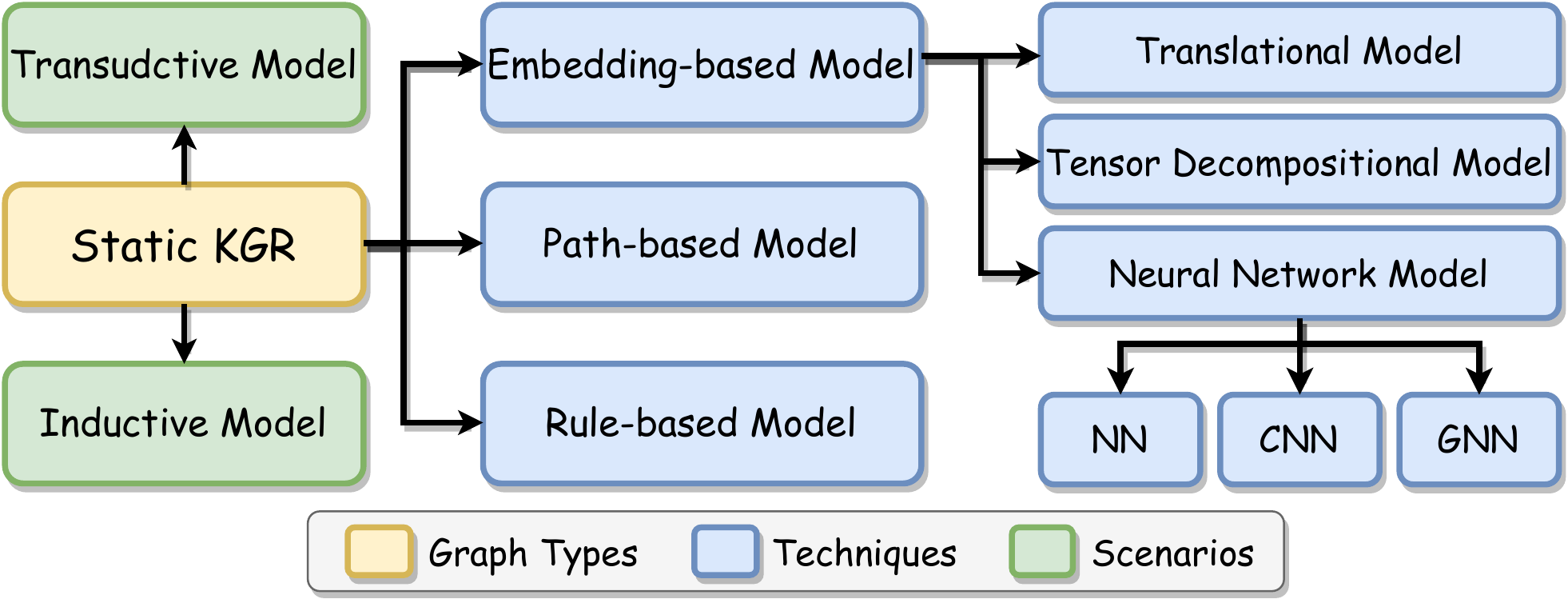}
\caption{Taxonomy of the static KGR models.}
\label{StaticTax}
\vspace{-0.3 cm}
\end{figure}

\begin{figure}[t]
\centering
\includegraphics[width=0.48\textwidth]{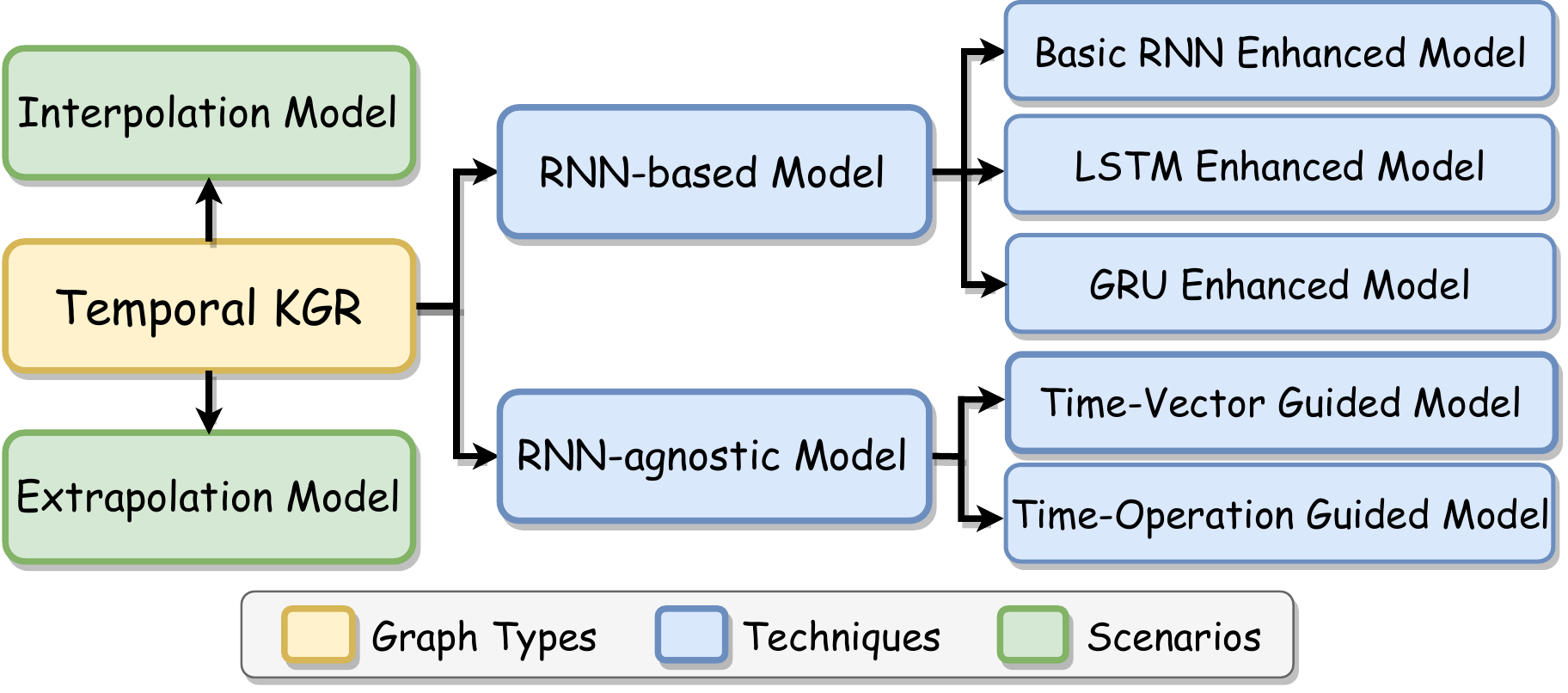}
\caption{Taxonomy of the temporal KGR models.}
\label{TemTax}  
\vspace{-0.15 cm}
\end{figure}

\begin{figure}[t]
\centering
\includegraphics[width=0.42\textwidth]{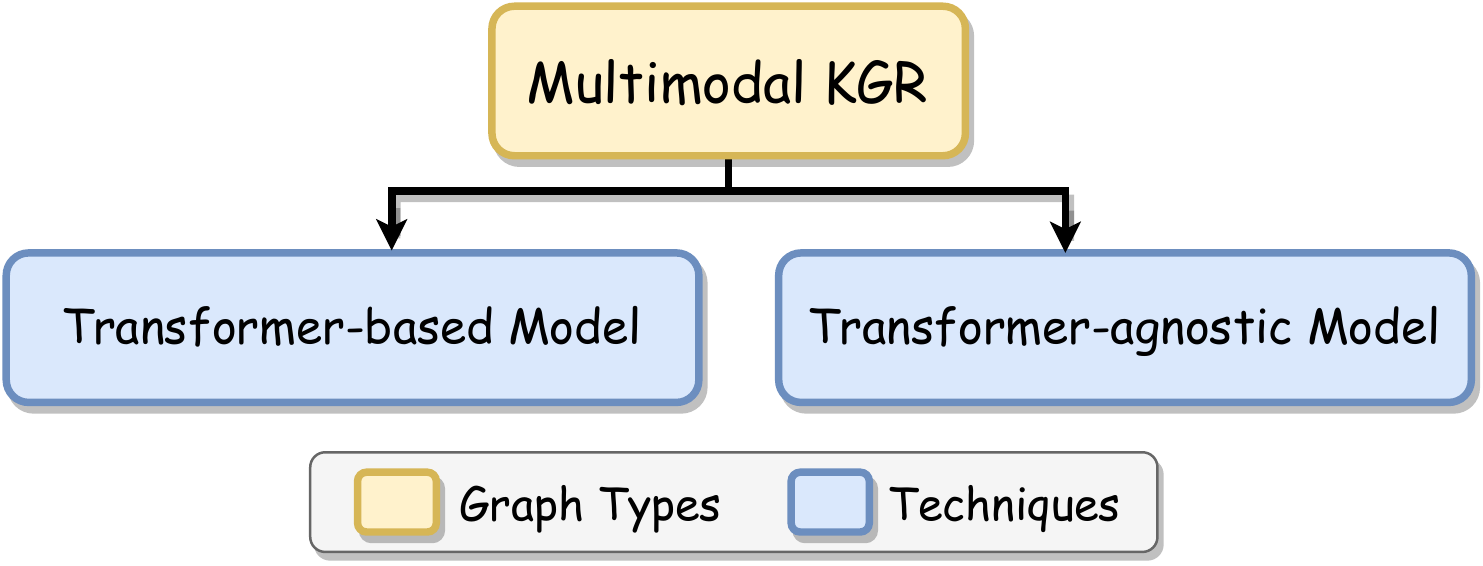}
\caption{Taxonomy of the multi-modal KGR models.}
\vspace{-0.25 cm}
\label{multitax}  
\end{figure}

We design a bi-level taxonomy to review existing KGR models for different KGs systematically. Specifically, three taxonomy criterion is adopted to classify the reviewed KGR models, \textit{i.e.,} graph types, techniques, and scenarios. As shown in Figure \ref{tax}, Graph types are the top-level taxonomy, which contains three KG types, \textit{i.e.,} static, temporal, and multi-modal KGs. Additionally, the base-level taxonomy consists of techniques (fourteen types) and scenarios (four types). Given a KGR model, we will first categorize it based on the top-level taxonomy. Then, we will further classify it according to its specific techniques and scenarios. As mentioned before, KGR, over different graph types, focus on different techniques and scenarios. Concretely, (1) as for techniques, embedding-based (five sub-types), path-based, and rule-based models are typical techniques for static KGR (See Figure \ref{StaticTax}). Besides, RNN-based (three sub-types) and RNN-agnostic (two sub-types) models are the designed technique criterion for temporal KGR models (See Figure \ref{TemTax}). Moreover, Transformer-based and Transformer-agnostic are the adopted technique criterion for multi-modal KGR models (See Figure \ref{multitax}). (2) As for reasoning scenarios, we only discuss this taxonomy in static and temporal KGR, \textit{i.e.,} transductive and inductive scenarios for static KGR, interpolation and extrapolation scenarios for temporal KGR.

\section{Static KGR Model}
We systematically introduce \textbf{90} static KGR models based on techniques and scenarios (See Table \ref{SUM_SKGR}).\vspace{-0.3cm}

\subsection{Review on Reasoning Techniques}
Static KGR models can be categorized into embedding-based, path-based, and rule-based models. Details are described below.

\subsubsection{Embedding-based Model}
\begin{figure*}[t]
\centering
\setlength{\abovecaptionskip}{-0.05cm}
\includegraphics[width=0.96\textwidth]{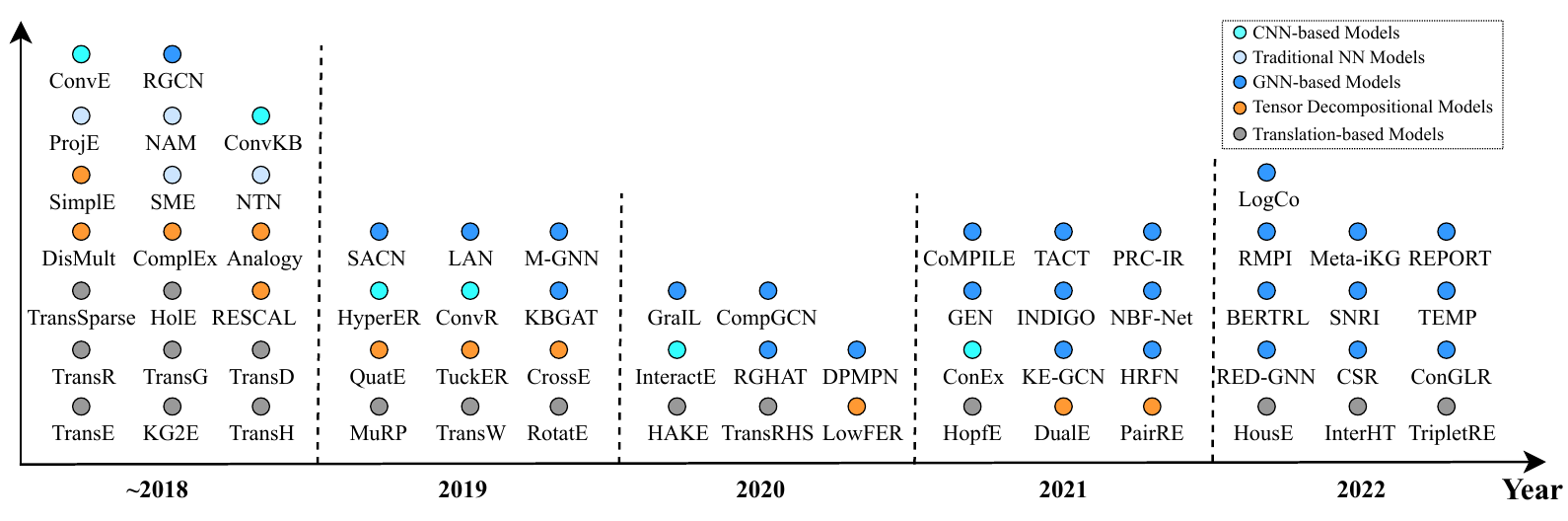}
\vspace{0.2 cm}
\caption{Timeline of the embedding-based models for static KGR.}
\label{Time-Embed}  
\vspace{-0.3 cm}
\end{figure*}

Embedding-based models learn the embedding vectors based on existing fact triplets and then rank top $k$ candidate facts based on the likelihood calculated by scoring functions. In general, there are three types, \textit{i.e.,} translational, tensor decompositional, and neural network models. Due to the majority quantity, the timeline of embedding-based models is present in Figure \ref{Time-Embed} for clear presentation.

\paragraph{Translational Model} Translational models regard relation $r$ as translational transformation to project entity $\textbf{e}$ into the latent space.

TransE \cite{TransE}, as the first translational model, regards the relation as a simple translation operation, \textbf{$e_h$}+\textbf{$r$} $\approx$ \textbf{$e_t$}. Although proven effective, it cannot handle some specific relations, such as one-to-many, many-to-one, symmetric and transitive relations. To address these limitations, lots of translational models are developed. The entities are encoded into the relation-specific hyperplane in TransH \cite{TransH}, which achieves better reasoning performance on one-to-many, many-to-one relations. Besides, TransR \cite{TransR} leverages distinct latent spaces for entities and relations and gets better expressive ability on transitive relation reasoning. Moreover, TransD \cite{TransD} first considers the scalability issues by leveraging independent projection vectors for entities and relations for large KGs. Afterward, probabilistic principles are integrated to model the uncertainty in KGR. For example, KG2E \cite{KG2E} leverages the Gaussian distribution covariance, and TransG \cite{TransG} makes use of the Bayesian technique for one-to-many relational facts. Meanwhile, to alleviate heterogeneity and imbalance issues, TranSparse \cite{TransSparse} provides an efficient solution by designing adaptive transfer sparser matrices, thus leading to better expressive ability. After that, TorusE \cite{TorusE} projects embeddings in a compact Lie group torus, and MuRP \cite{MuPR} designs a Möbius matrix-vector multiplication and Möbius addition for entity embedding projection, which all show better accuracy and scalability. Meanwhile, TransW \cite{TransW} first has enriched the entity and relation embedding with the word embeddings, achieving better performance in inferring facts with unseen entities or relations. Moreover, RotatE \cite{RotatE} proposes a rotation-based translational method with complex-valued embeddings to better infer the symmetry, anti-symmetry, inversion, and composition facts. Besides, HAKE \cite{HAKE} models the semantic hierarchy rather than relation patterns based on the polar coordinate space. Then, TransRHS \cite{TransRHS} first considers the Relation Hierarchical Structure (RHS) by incorporating RHS seamlessly into the embeddings. Besides, to handle the complex relational facts with a unified model, PairRE \cite{PairRE} models each relation representation with paired vectors to adaptive adjustment for complex relations, and HousE \cite{HousE} involves a novel parameterization based on the designed Householder transformations for rotation and projection. Nowadays, there are also some interesting attempts for translational models for more sufficient interactions, such as TripleRE \cite{TripleRE} and InterHT \cite{wang2022interht}. TripleRE creatively divides the relationship vector into three parts, takes advantage of the concept of residual, and achieves better performance. While, InterHT enhances the information interactions between the tail and head, improving the model capacity.

\paragraph{Tensor Decompositional Model}
Tensor decompositional models encode the KGs as three-way tensors, decomposed into a combination of low-dimensional vectors for entities and relations. 

As the first tensor decompositional model, RESCAL \cite{RESCAL} captures the latent semantics of each entity with vectors and further leverages the matrix to model the pairwise interactions among latent factors as a matrix. However, the model is complex with $O(d^2)$ parameters. To simplify it, DistMult \cite{DisMult} uses bi-linear diagonal matrices to reduce parameters to $O(d)$ per relation. Then, ComplEx \cite{ComplEx} generalizes DistMult by using complex-valued embeddings, which improve asymmetric relations modeling. Meanwhile, HolE \cite{HolE} models the holographic reduced representations and circular correlation, and Analogy \cite{Analogy} designs the bi-linear scoring function with analogical structural constraints for analogical reasoning, which both try to capture rich interactions between entities. Then, some models start to substitute the decompositional operations. SimplE \cite{SimplE} enhances the Canonical Polyadic (CP) decompositional for two independent entity embeddings, and Tucker Decomposition is first used by Tucker \cite{TuckER}. Meanwhile, CrossE \cite{CrossE} considers crossover interactions between entities via a relation-specific interaction matrix. QuatE \cite{QuatE} empowers the semantic matching between head and tail based on relational rotation quaternion representations. Inspired by it, DualE \cite{DualE} projects the embeddings in dual quaternion space to achieve a unified framework for both translation and rotation operations. Besides, HopfE \cite{HopfE} makes use of both structural and semantic attributes in 4D hyper-sphere space without losing interpretability. Besides expressive ability, efficiency also draws increasing attention these years. A factorized bi-linear pooling model is proposed based on Tucker decomposition, termed LowFER \cite{LowFER}, which is more efficient and lightweight. Moreover, QuatRE \cite{QuatRE} learns the quaternion embeddings with two novel operations, \textit{i.e.,} enhancing the correlations between entities with Hamilton product for entity embeddings and reducing computation by simplifying the translation matrices.

\paragraph{Neural Network Model}
Neural network (NN) models have yielded remarkable performance for KG reasoning these years. Three sub-types are defined based on the NN techniques, \textit{i.e.,} traditional neural network (NN), convolutional neural network (CNN), and graph neural network (GNN) models.

\emph{1) Traditional NN Model}: SME \cite{SME} first encodes the entities and relations into the latent space using neural networks. Meanwhile, neural tensor networks are used in NTN \cite{NTN} for relation reasoning in KGs. Then, NAM \cite{NAM} proposes the relational-modulated neural network (RMNN), and NN models shares variables in ProjE \cite{ProjE}, which jointly learns embeddings of the entities and relations via the standard loss function. These traditional NN models show great potential on static KGR while they suffer from learning shallow and less expressive features. 

\emph{2) CNN Model:} To learn deeper features, convolutional neural networks (CNNs) are integrated with KGR models. ConvE \cite{ConvE} first leverages 2D convolutional layers for KGR. ConvKB \cite{ConvKB} extends ConvE by removing the reshaping operation and captures global and transitional characteristics within facts for informative expression. Later on, HypER \cite{HypER} uses fully connected layers and relation-specific convolutional filters for better performance. Besides, ConvR \cite{convR} designs an adaptive convolutional network designed to maximize entity-relation interactions by constructing convolution filters across the entity and relation representations. Moreover, novel operations, \textit{i.e.,} feature reshaping, feature permutation, and circular convolution, are designed in InteractE \cite{InteractE} to handle complex interactions. Meanwhile, ConEx \cite{ConEx} integrates the affine transformation and a Hermitian inner product on complex-valued embeddings with the convolutional operation, which shows good expressiveness. CNN models generally perform better than traditional NN models. However, the information underlying graph structures cannot be well learned. 

\emph{3) GNN Model:} Graph neural networks, which are widely used for graph tasks, are also rapidly applied to KG reasoning. RGCN \cite{RGCN} uses the relation-specific transformation to aggregate neighborhood information. Then, each entity is encoded into a vector, and the decoder \textit{i.e.,} scoring function, reconstructs the facts based on entity representations. While RGCN omits the variances of entities, which hinders expressive ability. To alleviate it, attention mechanisms are integrated into lots of models, such as M-GNN \cite{M-GNN},  KBGAT \cite{KBGAT}, and \cite{LAN}. In particular, KBGAT \cite{KBGAT} leverages attention-based feature embeddings for better reasoning performance. Meanwhile, SACN \cite{SACN} leverages the weighted graph convolutional network (WGCN) as the encoder and a convolutional network called Conv-TransE as the decoder, which is effective. Afterward, TransGCN \cite{TransGCN} trains both relation and entity embeddings simultaneously with the transformation operator for relations. Later on, DPMPN \cite{DPMPN} and RGHAT \cite{RGHAT} designs a two-GNN framework to simultaneously encode information in different levels separately, \textit{i.e.,} global $\&$ local information for DPMPN and relation $\&$ entity information for RGHAT. After that, KE-GCN \cite{KE-GCN} jointly propagates and updates the embedding of both entities and edges. Similarly, COMPGCN \cite{COMPGCN} also jointly learns the representations with various entity-relation composition operations. Recently, more and more researchers have tried to handle out-of-knowledge-graph scenarios. GEN \cite{GEN} and HRFN \cite{HRFN} learn entity embeddings based on meta-learning for both seen-to-unseen and unseen-to-unseen facts. INDIGO \cite{INDIGO} is then proposed based on a GNN using pair-wise encoding. Besides, the GraIL-based model is a group of typical GNN models for inductive scenarios. The prototype GraIL \cite{GraIL}, as the landmark GNN-based model, first leverages RGCN to perform the reasoning based on the local enclosing subgraph. Based on it, many incremental works are developed, including TACT \cite{TACT}, CoMPILE \cite{CoMPILE}, Meta-iKG \cite{Meta-iKG}, SNRI \cite{SNRI}, RPC-IR \cite{PRC-IR}, and etc. These GraIL-based models all achieve promising inductive performances. Among them, TACT \cite{TACT} and CoMPILE \cite{CoMPILE} both raise the importance of relation embeddings in the task. Concretely, TACT \cite{TACT} uses topology-aware correlations between relations to generate representations for triplet scoring, which also inspires RMPI\cite{RMPI} and TEMP \cite{TEMP}. Besides, CoMPILE enhances the message interactions between relations and entities with a novel mechanism. After that, motivated by the great success of contrastive mechanisms \cite{simclr, KGESymCL}, contrastive learning models have been increasingly proposed, \textit{e.g.,} RPC-IR \cite{PRC-IR}, SNRI \cite{SNRI} etc. Besides, Meta-iKG \cite{Meta-iKG} verifies the effectiveness of meta-learning in the KGR task. After that, researchers try to make the reasoning more efficient. NBF-net \cite{NBF-net} and RED-GNN \cite{RED-GNN} achieve better efficiency by leveraging the traditional algorithm \textit{i.e.,} bellman-ford algorithm and dynamic programming to optimize the propagation strategy in previous GNN models. Besides, pGAT \cite{pGAT} leverages the EM algorithm for efficient learning. Moreover, BERTRL \cite{BERTRL} and ConGLR \cite{ConGRL} integrates the context for each entity to enhance the reasoning on the KGs. In particular, BERTRL \cite{BERTRL} can handle unseen relational facts. Regarding it, CSR \cite{CSR} deeply mines the logic rules underlying the structure patterns instead of the paths.
\begin{table*}[t]
\fontsize{5.5}{6.5}\selectfont 
\caption{Summary of the static knowledge graph reasoning models.}
\vspace{-0.25 cm}
\resizebox{\linewidth}{!}{
\begin{tabular}{ccccc|cccc}\hline
\textbf{Year} & \textbf{Model} & \textbf{Scenario} & \textbf{Technique}     & \textbf{} & \textbf{Year} & \textbf{Model} & \textbf{Scenario} & \textbf{Technique}     \\
\hline
2022          & LogCo \cite{LogCo}        & Inductive                  & GNN                    &           &2019          & M-GNN  \cite{M-GNN}        & Transductive               & GNN                                  \\
2022          & REPORT \cite{REPORT}        & Inductive                  & GNN                    &           & 2019          & SACN \cite{SACN}          & Transductive               & GNN                    \\
2022          & RED-GNN \cite{RED-GNN}        & Inductive                  & GNN                  &           & 2019          & KBGAT  \cite{KBGAT}        & Transductive               & GNN                    \\
2022          & ConGLR \cite{ConGRL}        & Inductive                  & GNN                        &           & 2019          & LAN    \cite{LAN}        & Inductive                  & GNN                    \\
2022          & TripleRE  \cite{TripleRE}     & Transductive               & Translational          &           & 2019          & CPL \cite{CPL}           & Transductive               & Relation Path          \\
2022          & InterHT \cite{wang2022interht}       & Transductive               & Translational          &           & 2019          & IterE \cite{IterE}         & Inductive                  & Logic Rule             \\
2022          & HousE \cite{HousE}         & Transductive               & Translational               &           & 2019          & pLogicNet \cite{pLogicNet}      & Inductive                  & Logic Rule             \\
2022          & BERTRL \cite{BERTRL}        & Inductive                  & GNN                      &           & 2019          & DRUM \cite{DRUM}          & Inductive                  & Logic Rule             \\
2022          & SNRI  \cite{SNRI}         & Inductive                  & GNN                       &           & 2019          & RLvLR \cite{RLvLR}         & Inductive                  & Logic Rule             \\
2022          & TEMP \cite{TEMP}          & Inductive                  & GNN                       &           & 2019          & Neural-Num-LP \cite{Neural-Num-LP} & Inductive               & Logic Rule             \\
2022          & RMPI \cite{RMPI}          & Inductive                  & GNN                      &           & 2018          & SimplE \cite{SimplE}        & Transductive               & Tensor Decompositional \\
2022          & Meta-iKG \cite{Meta-iKG}      & Inductive                  & GNN                           &           & 2018          & ConvKB  \cite{ConvKB}       & Transductive               & CNN                    \\
2022          & CSR  \cite{CSR}          & Inductive                  & GNN       &           & 2018          & ConvE  \cite{ConvE}        & Transductive               & CNN                    \\
2022          & CURL  \cite{CURL}         & Transductive               & Relation Path           &           & 2018          & RGCN  \cite{RGCN}         & Transductive               & GNN                    \\
2022          & GCR   \cite{GCR}         & Inductive                  & Logic Rule            &           & 2018          & M-walk  \cite{M-walk}       & Transductive               & Relation Path          \\
2021          & PairRE \cite{PairRE}        & Transductive               & Translational   &           & 2018          & MultiHop \cite{MultiHop}       & Transductive               & Relation Path          \\
2021          & HopfE  \cite{HopfE}        & Transductive               & Tensor Decompositional &           & 2018          & DIVA \cite{DIVA}          & Transductive               & Logic Rule             \\
2021          & DualE  \cite{DualE}        & Transductive               & Tensor Decompositional           &           & 2018          & RuleN \cite{RuleN}         & Inductive                  & Logic Rule             \\
2021          & ConEx \cite{ConEx}         & Transductive               & CNN                        &           & 2018          & RUGE  \cite{RUGE}         & Inductive                  & Logic Rule             \\
2021          & KE-GCN  \cite{KE-GCN}       & Transductive               & GNN                     &           & 2017          & ANALOGY  \cite{Analogy}       & Transductive               & Tensor Decompositional \\
2021          & HRFN \cite{HRFN}          & Inductive                  & GNN              &           & 2017          & ProjE  \cite{ProjE}        & Transductive               & Traditional NN         \\
2021          & GEN  \cite{GEN}          & Inductive                  & GNN                          &           & 2017          & MINERVA  \cite{MINERVA}      & Transductive               & Relation Path          \\
2021          & INDIGO  \cite{INDIGO}       & Inductive                  & GNN                   &           & 2017          & DeepPath \cite{DeepPath}      & Transductive               & Relation Path          \\
2021          & NBF-Net  \cite{NBF-net}      & Inductive                  & GNN                        &           & 2017          & NTP \cite{NTP}           & Inductive                  & Logic Rule             \\
2021          & CoMPILE \cite{CoMPILE}       & Inductive                  & GNN                        &           & 2017          & NeuralLP \cite{NeuralLP}       & Inductive                  & Logic Rule             \\
2021          & TACT  \cite{TACT}         & Inductive                  & GNN                    &           & 2016          & TranSparse \cite{TransSparse}    & Transductive               & Translational          \\
2021          & RPC-IR  \cite{PRC-IR}       & Inductive                  & GNN                 &           & 2016          & TransG \cite{TransG}        & Transductive               & Translational          \\
2020          & HAKE  \cite{HAKE}         & Transductive               & Translational            &           & 2016          & HolE  \cite{HolE}         & Transductive               & Tensor Decompositional \\
2020          & TransRHS \cite{TransRHS}      & Transductive               & Translational    &           & 2016          & ComplEx  \cite{ComplEx}      & Transductive               & Tensor Decompositional \\
2020          & LowFER  \cite{LowFER}       & Transductive               & Tensor Decompositional               &           & 2016          & NAM  \cite{NAM}          & Transductive               & Traditional NN         \\
2020          & InteractE \cite{InteractE}     & Transductive               & CNN                      &           & 2016          & LogSumExp \cite{LogSumExp}     & Transductive               & Relation Path          \\
2020          & DPMPN \cite{DPMPN}         & Transductive               & GNN               &           & 2016          & KALE \cite{KALE}          & Inductive                  & Logic Rule             \\
2020          & RGHAT \cite{RGHAT}         & Transductive               & GNN              &           & 2015          & TransD \cite{TransD}        & Transductive               & Translational          \\
2020          & COMPGCN  \cite{COMPGCN}      & Transductive               & GNN                        &           & 2015          & TransR \cite{TransR}        & Transductive               & Translational          \\
2020          & GraIL  \cite{GraIL}        & Inductive                  & GNN                          &           & 2015          & KG2E \cite{KG2E}          & Transductive               & Translational          \\
2020          & ExpressGNN  \cite{ExpressGNN}   & Inductive                  & Logic Rule                     &           & 2015          & DISTMULT \cite{DisMult}      & Transductive               & Tensor Decompositional \\
2020          & pGAT \cite{pGAT}          & Inductive                  & GNN                    &           & 2015          & RNNPRA  \cite{RNNPRA}       & Transductive               & Relation Path          \\
2019          & RotatE  \cite{RotatE}       & Transductive               & Translational             &           & 2014          & TransH \cite{TransH}        & Transductive               & Translational          \\
2019          & TransW \cite{TransW}        & Inductive                  & Translational             &           & 2014          & ProPPR  \cite{ProPPR}       & Transductive               & Relation Path          \\
2019          & MuRP   \cite{MuPR}        & Transductive               & Translational     &           & 2013          & AMIE  \cite{AMIE}         & Inductive                  & Logic Rule            \\
2019          & QuatE  \cite{QuatE}        & Transductive               & Tensor Decompositional &           & 2013          & SME  \cite{SME}          & Transductive               & Traditional NN         \\
2019          & TuckER  \cite{TuckER}       & Transductive               & Tensor Decompositional &           & 2013          & NTN  \cite{NTN}          & Transductive               & Traditional NN         \\
2019          & CrossE \cite{CrossE}         & Transductive               & Tensor Decompositional                  &           & 2013          & TransE \cite{TransE}        & Transductive               & Translational         \\
2019          & ConvR \cite{convR}         & Transductive               & CNN                    &           & 2011          & RESCAL \cite{RESCAL}        & Transductive               & Tensor Decompositional \\
2019          & HypER \cite{HypER}         & Transductive               & CNN                         &           & 2010          & PRA \cite{PRA}           & Transductive               & Relation Path    \\
\hline  
\end{tabular}
}
\vspace{-0.5cm}
\label{SUM_SKGR}
\end{table*}
\begin{table}[!h]
\fontsize{10}{11}
\selectfont 
\caption{Performance comparison of static KGR models on WN18RR, FB15k-237 in the transductive scenario. Best results are marked as boldfaced. "H" is short for "Hits".}
\vspace{-0.2 cm}
\resizebox{\linewidth}{!}{
\begin{tabular}{ccccccccc}
\hline
\multicolumn{1}{c}{\multirow{2}{*}{\textbf{Methods}}} & \multicolumn{4}{c}{\multirow{1}{*}{\textbf{WN18RR}}}                                                                                                                & \multicolumn{4}{c}{\multirow{1}{*}{\textbf{FB15K-237}}}                                                                                                                                                                               \\
 \cline{2-9} 
\multicolumn{1}{c}{}                         & \multirow{1}{*}{\textbf{MRR}}               & \multirow{1}{*}{\textbf{H@1}}             & \multirow{1}{*}{\textbf{H@3}}             & \multicolumn{1}{c}{\multirow{1}{*}{\textbf{H@10}}} & \multirow{1}{*}{\textbf{MRR}}               & \multicolumn{1}{c}{\multirow{1}{*}{\textbf{H@1}}} & \multicolumn{1}{c}{\multirow{1}{*}{\textbf{H@3}}} & \multicolumn{1}{c}{\multirow{1}{*}\textbf{{H@10}}}     \\ \hline
\multicolumn{1}{c}{TransE \cite{TransE}}                   & 0.231                & 0.021                  & 0.409                  & \multicolumn{1}{c}{0.533}                   & 0.289                     & 0.193                     & 0.326                     & \multicolumn{1}{c}{0.478}                  \\
\multicolumn{1}{c}{RotatE \cite{RotatE}}                   & 0.476                & 0.428                  & 0.492                  & \multicolumn{1}{c}{0.571}                   & 0.338                     & 0.241                     & 0.375                     & \multicolumn{1}{c}{0.533}                   \\
\multicolumn{1}{c}{QuatE \cite{QuatE}}                    & 0.481                & 0.436                  & 0.500                    & \multicolumn{1}{c}{0.564}                   & 0.311                     & 0.221                     & 0.342                     & \multicolumn{1}{c}{0.495}                     \\
InteractE  \cite{InteractE}  & 0.463 & 0.430 & --     & 0.528 & 0.354 & 0.263 & --     & 0.535  \\
\multicolumn{1}{c}{DualE \cite{DualE}}                    & 0.482                & 0.440                   & 0.500                    & \multicolumn{1}{c}{0.561}                   & 0.330                     & 0.237                     & 0.363                     & \multicolumn{1}{c}{0.518}                                \\
\multicolumn{1}{c}{HAKE  \cite{HAKE}}                     & {{0.497}}       & {0.453}         & {{0.515}}         & \multicolumn{1}{c}{0.582}                   & 0.335                     & 0.237                     & 0.371                     & \multicolumn{1}{c}{0.530}                   \\ 
MuRP \cite{MuPR}     & 0.481 & 0.440 & 0.495 & 0.566 & 0.335 & 0.243 & 0.367 & 0.518    \\
ConEx  \cite{ConEx}   & 0.481 & 0.448  & 0.493 & 0.550  & 0.366 & 0.271 & 0.403 & 0.555 \\
\multicolumn{1}{c}{HousE \cite{HousE}}                     & {\textbf{0.511}}       & {0.465}         & {\textbf{0.528}}         & \multicolumn{1}{c}{0.602}                   & 0.361                     & 0.266                     & 0.399                     & \multicolumn{1}{c}{0.551}                             \\ 
\multicolumn{1}{c}{RESCAL \cite{RESCAL}}                   & 0.455                & 0.419                  & 0.461                  & \multicolumn{1}{c}{0.493}                   & \multicolumn{1}{c}{0.353} & \multicolumn{1}{c}{0.264} & \multicolumn{1}{c}{0.385} & \multicolumn{1}{c}{0.528}                    \\
\multicolumn{1}{c}{DisMult \cite{DisMult}}                  & 0.420                & 0.370                  & 0.439                  & \multicolumn{1}{c}{0.521}                  & 0.243                     & 0.191                     & 0.271                     & \multicolumn{1}{c}{0.328}                   \\
\multicolumn{1}{c}{ComplEX \cite{ComplEx}}                  & 0.440                & 0.410                  & 0.460                  & \multicolumn{1}{c}{0.510}                   & \multicolumn{1}{c}{0.346} & \multicolumn{1}{c}{0.256} & \multicolumn{1}{c}{0.386} & \multicolumn{1}{c}{0.525}                         \\
TuckER \cite{TuckER}    & 0.470  & 0.443 & 0.482 & 0.526 & 0.358 & 0.266 & 0.394 & 0.544  \\
\multicolumn{1}{c}{ConvE \cite{ConvE}}                    & 0.430                & 0.400                  & 0.440                  & \multicolumn{1}{c}{0.520}                   & \multicolumn{1}{c}{0.325} & \multicolumn{1}{c}{0.237} & \multicolumn{1}{c}{0.356} & \multicolumn{1}{c}{0.501}                                            \\
HypER \cite{HypER}    & 0.465 & 0.443 & 0.477 & 0.522 & 0.341 & 0.252 & 0.376 & 0.520   \\
ConvKB  \cite{ConvKB}  & 0.265 & 0.058 & 0.445 & 0.558 & 0.289 & 0.198 & 0.324 & 0.471   \\
ConvR  \cite{convR}   & 0.475 & 0.443 & 0.489 & 0.537 & 0.350  & 0.261 & 0.385 & 0.528  \\
\multicolumn{1}{c}{ComplEX-DURA \cite{ComplEx}}             & 0.489                & 0.445                  & 0.503                  & \multicolumn{1}{c}{0.574}                   & \multicolumn{1}{c}{0.370} & \multicolumn{1}{c}{0.275} & \multicolumn{1}{c}{0.409} & \multicolumn{1}{c}{0.562}                        \\ 
LowFER \cite{LowFER}   & 0.465 & 0.434 & 0.479 & 0.526 & 0.359 & 0.359 & 0.266 & 0.396  \\
\multicolumn{1}{c}{RGCN \cite{RGCN}}                     & 0.427                & 0.382                  & 0.446                  & \multicolumn{1}{c}{0.510}                   & 0.248                     & 0.153                     & 0.258                     & \multicolumn{1}{c}{0.414}                \\
\multicolumn{1}{c}{SACN \cite{SACN}}                     & 0.470                & 0.430                  & 0.480                  & \multicolumn{1}{c}{0.540}                   & \multicolumn{1}{c}{0.350} & \multicolumn{1}{c}{0.260} & \multicolumn{1}{c}{0.390} & \multicolumn{1}{c}{0.540}                                  \\
\multicolumn{1}{c}{KBGAT \cite{KBGAT}}                    & 0.464                & 0.426                  & 0.479                  & \multicolumn{1}{c}{0.539}                   & \multicolumn{1}{c}{0.350} & \multicolumn{1}{c}{0.260} & \multicolumn{1}{c}{0.385} & \multicolumn{1}{c}{0.531}                   \\
\multicolumn{1}{c}{COMPGCN \cite{COMPGCN}}                  & 0.469                & 0.434                  & 0.482                  & \multicolumn{1}{c}{0.537}                   & 0.352                     & 0.261                     & 0.387                     & \multicolumn{1}{c}{0.534}                                 \\ 
DPMPN \cite{DPMPN}    & 0.482 & 0.444 & 0.497 & 0.558 & 0.369 & 0.286 & 0.403 & 0.530  \\
RGHAT  \cite{RGHAT}   & 0.483 & 0.425 & 0.499 & 0.588 & \textbf{0.522} & \textbf{0.462} & \textbf{0.546} & \textbf{0.631} \\
\multicolumn{1}{c}{RED-GNN \cite{RED-GNN}}                  & 0.533               & \textbf{0.485}                  &         --       & \multicolumn{1}{c}{0.624}                   & 0.374                     & 0.283                     &--                   & \multicolumn{1}{c}{0.558}                      \\
NeuralLP \cite{NeuralLP} & 0.459 & {0.376} & 0.468 & \textbf{0.657} & 0.227 & 0.166 & 0.248 & 0.348 \\
MINERVA \cite{MINERVA}  & 0.448 & 0.413 & 0.456 & 0.513 & 0.271 & 0.192 & 0.307 & 0.426 \\
M-walk  \cite{M-walk}  & 0.437 & 0.415 & 0.447 & 0.543 & 0.234 & 0.168 & 0.245 & 0.403  \\
pLogicNet \cite{pLogicNet}& 0.441 & 0.398 & 0.446 & 0.537 & 0.332 & 0.237 & 0.367 & 0.524 \\
\multicolumn{1}{c}{CURL \cite{CURL}}                  & 0.460               & 0.429                  &           0.471        & \multicolumn{1}{c}{0.523}                   & 0.306                     & 0.224                     & 0.341                    & \multicolumn{1}{c}{0.470}                   \\
\hline
\end{tabular}}
\vspace{-0.2 cm}
\label{STPR}
\end{table}

\begin{table*}[t]
\fontsize{4}{4.6}\selectfont 
\caption{Hits@10 Performance comparison (in percentage) of static KGR models on WN18RR, FB15k-237 and NELL-995 in the inductive scenario. Best results are marked as boldfaced.}
\vspace{-0.2 cm}
\resizebox{\linewidth}{!}{
\begin{tabular}{ccccccccccccccc}
\hline
\multirow{2}{*}{\textbf{Methods}}                    & \multicolumn{4}{c}{\textbf{WN18RR}}    &  \multicolumn{4}{c}{\textbf{FB15K-237}} &  \multicolumn{4}{c}{\textbf{NELL-995}}  \\ \cline{2-5} \cline{6-9} \cline{10-13} 
& \textbf{v1}    & \textbf{v2}    & \textbf{v3}    & \textbf{v4}    &  \textbf{v1}    & \textbf{v2}    & \textbf{v3}    & \textbf{v4}   &  \textbf{v1}    & \textbf{v2}    & \textbf{v3}    & \textbf{v4}    \\ \hline
\multicolumn{1}{c}{Neural-LP \cite{NeuralLP}} & 74.37 & 68.93 & 46.18 & 67.13 &  52.92 & 58.94 & 52.90 & 55.88 &  40.78 & 78.73 & 82.71 & 80.58 \\
\multicolumn{1}{c}{DRUM \cite{DRUM}}      & 74.37 & 68.93 & 46.18 & 67.13 &  52.92 & 58.73 & 52.90 & 55.88 &  19.42 & 78.55 & 82.71 & 80.58 \\
\multicolumn{1}{c}{RuleN \cite{RuleN}}     & 80.85 & 78.23 & 53.39 & 71.59 &  49.76 & 77.82 & {87.69} & 85.60 & 53.50 & 81.75 & 77.26 & 61.35 \\ 
\multicolumn{1}{c}{GraIL \cite{GraIL}}     & 82.45 & 78.68 & 58.43 & 73.41 &   64.15 & 81.80 & 82.83 & {89.29} &  59.50 & 93.25 & 91.41 & 73.19 \\
\multicolumn{1}{c}{TACT \cite{TACT}}   & {83.24} & {{81.63}} &  {{62.73}} & {76.27} &  {65.61} &  {83.05} & {87.28} &  {91.33} &  {55.00} &  {88.97} &  {91.35} &  {74.69} \\			
\multicolumn{1}{c}{CoMPILE \cite{CoMPILE}}   & 83.60 & 79.82 & 60.69 & 75.49 &  {67.64} & {82.98} & {84.67} & 87.44 &  58.38 & \textbf{93.87} & 92.77 & {75.19} \\
 \multicolumn{1}{c}{Meta-iKG \cite{Meta-iKG}} & --     & --     & --     & --  &   66.52 & 72.37 & 68.81 & 74.32 & {60.49} & 74.07 & 77.99 & 71.63 \\
\multicolumn{1}{c}{RPC-IR \cite{PRC-IR}}    & {85.11} & {81.63} & 62.40 & {76.35} &  67.56 & 82.53 & 84.39 & {89.22} &  59.75 & {93.28} & {94.01} & 71.82 \\ 
\multicolumn{1}{c}{SNRI \cite{SNRI}}    & {87.23} & {83.10} & {67.31} & 83.32 & {71.79} & {86.50} & {89.59} & 89.39 & {--} & -- & {--}  & {--} \\
 RMPI  \cite{RMPI}  & 82.45 & 78.68 & 58.68 & 73.41 & 65.37 & 81.80  & 81.10  & 87.25 & 59.50  & 92.23 & 93.57 & \textbf{87.62} \\
REPORT \cite{REPORT} & 88.03 & 85.83 & \textbf{72.31} & 81.46 & 71.69 & \textbf{88.91} & \textbf{91.62} & \textbf{92.28} & {--} & -- & {--}  & {--}   \\
ConGLR \cite{ConGLR} & 85.64 & \textbf{92.93} & 70.74 & \textbf{92.90}  & 68.29 & 85.98 & 88.61 & 89.31 & {--} & -- & {--}  & {--}    \\
LogCo \cite{LogCo}  & \textbf{90.16} & 84.69 & {68.68} & 79.08 & \textbf{73.90}  & 81.91 & 80.64 & 84.20  & \textbf{61.75} & 93.48 & \textbf{94.19} & 80.82\\
 \hline
\end{tabular}
}
\label{SIPR}
\vspace{-0.1 cm}
\end{table*}

\subsubsection{Path-based Model}
Path-based models mine the logical knowledge underlying the paths between the queried head and tail to achieve reasoning.

Random walk \cite{rw1}  inferences have been widely investigated. For instance, the Path-Ranking Algorithm (PRA) \cite{PRA} derives the path-based logic rules under path constraints. ProPPR \cite{ProPPR} further introduces space similarity heuristics by incorporating textual content to alleviate the feature sparsity issue in PRA. Meanwhile, Neural multi-hop path-based models are also studied for better expressive ability. By iteratively using compositionality, RNNPRA \cite{RNNPRA} leverages RNN to compose the implications of relational paths for reasoning. LogSumExp \cite{LogSumExp} designs a logical composition method across all the elements with attention mechanisms for multiple reasoning. Then, a unified variational inference framework is proposed by DIVA \cite{DIVA}, which separates multi-hop reasoning into two steps, \textit{i.e.,} path-finding and path-reasoning.
Deep reinforcement learning (DRL) techniques, such as the Markov decision process (MDP), have recently been used to reformulate path-finding between entities as a sequential decision-making task. The designed reinforcement learning agent learns to find the reasoning paths according to entity interactions, and the corresponding policy gradient is utilized for training. Concretely, different fine-grained manners are employed by different models. 
For example, DeepPath \cite{DeepPath} applies DRL for relational path learning via the novel rewards and relational action spaces, improving both the models' performance and efficiency. Meanwhile, MINERVA \cite{MINERVA} takes path finding between entities as a sequential optimization problem by maximizing the expected reward \cite{KGESURVEY}, which excludes the target answer entity for more capable reasoning. After that, MultiHop \cite{MultiHop} designs a soft reward mechanism instead of only relying on binary rewards, as well as the dropout action, which enables more effective path exploration. Besides, Monte Carlo Tree Search (MCTS) is used by M-Walk \cite{M-walk} to generate the path, and CPL \cite{CPL} proposes collaborative policy learning for path-finding and fact extraction by leveraging the text corpus corresponding to the entities. Moreover, two agents in different levels, \textit{i.e.,} DWARF AGENT at the entity level and GIANT AGENT at the cluster level are proposed in CURL \cite{CURL}, which collaborate to achieve optimal reasoning performance.

\vspace{-0.1 cm}
\subsubsection{Rule-based Model}
Rule-based models aim to make better use of such symbolic features underlying the logic rule, which is generally defined in the form of B $\rightarrow$ A, where A is a fact, and B can be a set of facts.

Logical rules can be extracted from KG for reasoning by rule mining tools, \textit{e.g.,} AMIE \cite{AMIE}, RuleN \cite{RuleN}, etc. Then, a more scalable rule mining approach via the techniques of rule searching and pruning is designed by RLvLR \cite{RLvLR}. After that, 
how to inject logical rules into embeddings for better reasoning performance has drawn increasing research attention \cite{KGESURVEY}. In general, there are two ways for it, \textit{i.e.,} joint learning and iterative training.
For instance, KALE \cite{KALE} is a unified joint model by leveraging the t-norm fuzzy logical connectives between compatible facts and rule embedding.
Besides, RUGE \cite{RUGE} is an iterative model utilizing the soft rules for embedding rectification. Inspired by it, the iterative training strategy, composed of embedding learning, axiom induction, and axiom injection, is designed by IterE \cite{IterE}. After that, researchers integrate neural network techniques into the rule-based models to alleviate the issues of limited expressive ability and huge space consumption of the previous rule-based models. Neural Theorem Provers (NTP) \cite{NTP} mines logical rules with the designed radial kernel. Besides, NeuralLP \cite{NeuralLP} leverages attention mechanisms and auxiliary memory to optimize the gradients for mining the rules, and Neural-Num-LP \cite{Neural-Num-LP} further integrates the cumulative sum operations and dynamic programming with NeuralLP to learn numerical rules. Meanwhile, an end-to-end differentiable rule-based model is proposed in DRUM \cite{DRUM}. Then, the probabilistic logic neural network is designed in pLogicNet \cite{pLogicNet}, which shows great performances for first-order logic mining. ExpressGNN \cite{ExpressGNN} further generalizes it by finetuning GNN models for more efficient reasoning. Moreover, GCR \cite{GCR} achieves promising performance for both reasoning and recommendation by mining the neighborhood information around the queried facts.

\subsection{Review on Reasoning Scenarios}
Based on the observation, there are \textbf{56} transductive models and \textbf{34} inductive models (See Table \ref{SUM_SKGR}). Among them, 56.25\% of the inductive models belong to GNN models, and 37.5\% belongs to rule-based models. In particular, none of the path-based models has shown an incredible inductive ability for reasoning. Still, most of the rule-based models are good at inductive scenarios, which is reasonable. Since path-based models are developed based on searching for specific paths, the trained models in this manner are hardly applied when unseen elements occur. While most rule-based models can derive entity-agnostic logical rules from the KGs, and the invariance brought by the rules can be easily applied to inductive scenarios. Moreover, we present a fair performance comparison of the typical SOTA static KGR models for transductive and inductive scenarios separately (See Table \ref{STPR} and Table \ref{SIPR}). The results support the analysis above. Additionally, embedding-based models, especially for the GNN models, have shown good capacity and compatibility for both scenarios and also most of the recent research lies in this type.

\subsection{Observation and Discussion}
Based on the above reviews, we can get the following observations, which may indicate the scope for different static KGR models and reveal the future trend in static KGR. \textbf{(1)} Embedding-based models generally have the better expressive ability but lack explainability. Meanwhile, more attention is currently focused on developing GNN-based models since the KGR models require a high-quality representation of the relational facts and the graph structure, which is most suitable for GNN models. \textbf{(2)} Path-based and Rule-based models are more explainable than embedding-based models, but they usually suffer from limited expressive ability and huge complexity of time and space. \textbf{(3)} Most of the path-based models are more suitable for transductive reasoning due to the path-searching schemes, while Rule-based models naturally inherit the inductive ability due to the generalization of the rule paradigms. \textbf{(4)} For a long time, transductive reasoning models have kept appearing and greatly impacted both academic research and industrial applications. However, due to the issues of scalability and expressive ability issues, researchers have recently focused on developing inductive reasoning models. As a brief conclusion, we would encourage to study more on GNN-based models, considered the most promising model type for static KGR.

\section{Temporal KGR Model}
We systematically introduce \textbf{58} temporal KGR models according to techniques, \textit{i.e.,} how they integrate time information and scenarios (See Table \ref{Sum_Tem}).\vspace{-0.2cm}

\subsection{Review on Reasoning Techniques}
Temporal KGR models can be categorized into RNN-based, and RNN-agnostic models. Details are described below.

\begin{table*}[t]
\fontsize{5}{6}\selectfont 
\caption{Summary of the temporal knowledge graph reasoning models.}
\vspace{-0.2 cm}
\resizebox{\linewidth}{!}{
\begin{tabular}{ccccc|cccc}
\hline
\textbf{Year} & \textbf{Model} & \textbf{Scenario} & \textbf{Technique}     & \textbf{} & \textbf{Year} & \textbf{Model} & \textbf{Scenario} & \textbf{Technique} \\
\hline
2023                              & RETIA \cite{RETIA}                              & Extrapolation                                  & GRU     &  & 2021                              & xERTE \cite{xERTE}                             & Extrapolation                                  & Time-Vector    \\
2023                              & RPC \cite{RPC}                              & Extrapolation                                  & GRU      &  & 2021                              & CyGNet \cite{CyGNet}                            & Extrapolation                                  & Time-Vector \\
2022                              & CENET \cite{CENET}                              & Extrapolation                                  & Time-Operation    &  & 2021                              & TIE   \cite{TIE}                             & Interpolation                                  & Time-Operation\\
2022                              & DA-Net \cite{DA-Net}                            & Extrapolation                                  & Time-Operation     &  & 2021                              & TeLM    \cite{TeLM}                           & Interpolation                                  & Time-Operation\\
2022                              & HiSMatch \cite{HiSMatch}                           & Extrapolation                                  & GRU      &  & 2021                              & ChronoR   \cite{ChronoR}                         & Interpolation                                  & Time-Vector \\
2022                              & rGalT \cite{rGalT}                             & Extrapolation                                  & Time-Operation    &  & 2021                              & RE-GCN  \cite{RE-GCN}                           & Extrapolation                                  & GRU        \\
2022                              & MetaTKGR \cite{MetaTKGR}                           & Extrapolation                                  & Time-Operation                      &  & 2021                              & RTFE    \cite{RTFE}                           & Interpolation                                  & Time-Operation        \\
2022                              & FILT  \cite{FILT}                             & Interpolation                                  & Time-Operation   &  & 2021                              & HIP     \cite{HIP}                           & Extrapolation                                  & GRU        \\
2022                              & TKGC-AGP \cite{TKGC-AGP}                           & Interpolation                                  & Time-Operation    &  & 2021                              & Tpath  \cite{Tpath}                            & Interpolation                                  & LSTM       \\
2022                              & Tlogic \cite{Tlogic}                            & Extrapolation                                  & Time-Operation    &  & 2020                              & TIMEPLEX   \cite{TIMEPLEX}                         & Interpolation                                  &Time-Operation  \\
2022                              & TLT-KGE \cite{TLT-KGE}                            & Interpolation                                  & Time-Vector    &  & 2020                              & DyERNIE   \cite{DyERNIE}                         & Interpolation                                  &Time-Operation  \\
2022                              & CEN \cite{CEN}                               & Extrapolation                                  & Time-Operation   &  & 2020                              & DacKGR   \cite{DacKGR}                          & Interpolation                                  & RNN   \\
2022                              & BoxTE \cite{BoxTE}                             & Interpolation                                  & Time-Vector  &  & 2020                              & TNTComplEx   \cite{TNTComplEx}                      & Interpolation                                  & Time-Vector     \\
2022                              & TempoQR \cite{TempoQR}                            & Interpolation                                  & Time-Vector & & 2020                              & TComplEx   \cite{TNTComplEx}                      & Interpolation                                  & Time-Vector     \\
2022                              & TuckERTNT     \cite{TuckERTNT}                     & Interpolation                                  & Time-Vector &  & 2020                              & TDGNN     \cite{TDGNN}                         & Extrapolation                                  & Time-Operation    \\
2022                              & GHT     \cite{GHT}                     & Extrapolation                                  & Time-Operation  &  & 2020                              & ATiSE   \cite{ATiSE}                           & Interpolation                                  & Time-Operation \\
2022                              & DKGE  \cite{DKGE}                             & Interpolation                                  & Time-Operation  &  & 2020                              & Diachronic   \cite{Diachronicembeddings}           & Interpolation                                  & Time-Operation \\
2022                              & TiRGN  \cite{TiRGN}                            & Extrapolation                                  & GRU        &  & 2020                              & DE-Simple   \cite{Diachronicembeddings}           & Interpolation                                  & Time-Operation \\
2022                              & RotateQVS     \cite{RotateQVS}                     & Interpolation                                  & Time-Vector        &  & 2020                              & TeRo    \cite{TeRo}                           & Interpolation                                  & Time-Operation\\
2022                              & ExKGR \cite{ExKGR}                             & Interpolation                                  & LSTM        &  & 2020                              & EvolveGCN    \cite{EvolveGCN}                      & Extrapolation                                  & LSTM+GRU        \\
2022                              & TRHyTE \cite{TRHyTE}                            & Interpolation                                  & GRU        &  & 2020                              & TeMP       \cite{TeMP2020}                        & Interpolation                                  & GRU        \\
2022                              & EvoKG  \cite{EvoKG}                            & Extrapolation                                  & RNN        &  & 2020                              & RE-NET     \cite{RE-NET}                        & Extrapolation                                  & RNN       \\
2021                              & TPmod  \cite{TPmod}                            & Interpolation                                  & GRU    &  & 2019                              & DyRep    \cite{DyRep}                          & Extrapolation                                  & Time-Operation\\
2021                              & TimeTraveler      \cite{TimeTraveler}                 & Extrapolation                                  & LSTM    &  & 2018                              & TTransE \cite{TTransE}                           & Interpolation                                  & LSTM    \\
2021                              & CluSTeR \cite{CluSTeR}                            & Extrapolation                                  & LSTM+GRU    &  & 2018                              & HyTE   \cite{HyTE}                            & Interpolation                                  &Time-Operation \\
2021                              & TPRec \cite{TPRec}                             & Extrapolation                                  & Time-Operation    &  & 2018                              & ChronoTranslate \cite{ChronoTranslate}                   & Interpolation                                  & Time-Operation  \\
2021                              & DBKGE \cite{DBKGE}                             & Interpolation                                  & Time-Vector    &  & 2018                              & TA-DISTMULT \cite{TA-DISTMULT}                       & Interpolation                                  & LSTM       \\
2021                              & TANGO \cite{TANGO}                             & Extrapolation                                  & Time-Operation    &  & 2018                              & TA-TransE \cite{TA-DISTMULT}                       & Interpolation                                  & LSTM       \\
2021                              & T-GAP \cite{T-GAP}                             & Interpolation                                  & Time-Vector   &  & 2017                              & Know-Evolve  \cite{Know-Evolve}                      & Extrapolation                                  & RNN  
\\
\hline
\end{tabular}}
\vspace{-0.1 cm}
\label{Sum_Tem}
\end{table*}

\subsubsection{RNN-based Model}
Recurrent Neural Networks (RNNs) are suitable for mining the changes over time. Thereby many temporal KGR models use RNNs to directly model the temporal information, termed RNN-based models. According to different variants of RNN, the models can be divided into three types, \textit{i.e.,} basic RNN enhanced models, LSTM enhanced models, and GRU enhanced models.

\paragraph{Basic RNN Enhanced Models} Some temporal KGR models can effectively model temporal information using basic RNN models. To name a few, Know-Evolve \cite{Know-Evolve} is a classical temporal KGR model that generates non-linearly entity embeddings over time. RE-NET \cite{RE-NET} applies the GCN and RNN models to capture the evolutional dynamics in temporal KGs to the query over time. EvoKG \cite{EvoKG} introduces the RNN model to mine the dynamic evolving structural information and models entity interactions by combining the neighborhood information.
\begin{table}[!t]
\fontsize{13}{16}\selectfont  
    \caption{Performance comparison (in percentage) of temporal KGR models on ICEWS14, ICEWS05-15 for interpolation scenario. Best results are marked as boldfaced. "H" is short for "Hits".}
  \vspace{-0.2cm}
  \centering
    \resizebox{\linewidth}{!}{
\begin{tabular}{ccccccccc}
\hline
\multirow{2}{*}{\textbf{Model}} & \multicolumn{4}{c}{\textbf{ICEWS05-15}} & \multicolumn{4}{c}{\textbf{ICEWS14}}   \\ \cline{2-9} 
& \textbf{MRR}    & \textbf{H@1}   & \textbf{H@3}   & \textbf{H@10}  & \textbf{MRR}    & \textbf{H@1}   & \textbf{H@3}   & \textbf{H@10}  \\ \hline
TA-TransE   \cite{TA-DISTMULT}               & 29.90  & 9.60  & --     & 66.80 & 27.50 & 9.50  & --    & 62.50 \\ 
TA-DISTMULT  \cite{TA-DISTMULT}            & 47.40  & 34.60 & --     & 72.80 & 47.70 & 36.30 & --     & 68.60 \\
HyTE    \cite{HyTE}                 & 31.60  & 11.60 & 44.50 & 68.10 & 29.70 & 10.80 & 41.60 & 65.50 \\
TTransE   \cite{TTransE}                & 27.10  & 8.40  & --     & 61.60 & 25.50 & 7.40  & --     & 60.10 \\
RE-NET   \cite{RE-NET}                & 43.70  & 33.60 & 48.80 & 62.72 & 39.86 & 30.11 & 44.02 & 58.21 \\
TeMP    \cite{TEMP}                 & 69.10  & 56.60 & \textbf{78.20} & \textbf{91.70} & 60.10 & 47.80 & 68.10 & \textbf{82.80} \\
TeRo    \cite{TeRo}                 & 58.60  & 46.90 & 66.80 & 79.50 & 56.20 & 46.80 & 62.10 & 73.20 \\
DE-SimplE    \cite{Diachronicembeddings}           & 51.30  & 39.20 & 57.80 & 74.80 & 52.60 & 41.80 & 59.20 & 72.50 \\
Diachronic   \cite{Diachronicembeddings}            & 51.30  & 39.20 & 57.80 & 74.80 & 52.60 & 41.80 & 59.20 & 72.50 \\
ATiSE  \cite{ATiSE}                     & 51.90  & 37.80 & 60.60 & 79.40 & 55.00 & 43.60 & 62.90 & 75.00 \\
TComplEx  \cite{TNTComplEx}              & 66.40  & 58.30 & 71.60 & 81.10 & 61.90 & 54.20 & 66.10 & 76.70 \\
TNTComplEx   \cite{TNTComplEx}              & 67.00  & 59.00 & 71.00 & 81.00 & 62.00 & 52.00 & 66.00 & 76.00 \\
DyERNIE  \cite{DyERNIE}                & \textbf{73.90}  & \textbf{67.90} & 77.30 & 85.50 & \textbf{66.90} & \textbf{59.90} & \textbf{71.40} & 79.70 \\
T-GAP    \cite{T-GAP}                 & 67.00  & 56.80 & 74.30 & 84.50 & 61.00 & 50.90 & 67.70 & 79.00 \\
TIMEPLEX   \cite{TIMEPLEX}     & 64.00  & 54.50 & --     & 81.80 & 60.40 & 51.50 & --     & 77.10 \\
RTFE   \cite{RTFE}              & 64.50  & 55.30 & 70.60 & 81.10 & 59.20 & 50.30 & 64.60 & 75.80 \\
ChronoR    \cite{ChronoR}                  & 68.41  & 61.06 & 73.01 & 82.13 & 62.53 & 54.67 & 66.88 & 77.31 \\
TeLM    \cite{TeLM}                    & 67.80  & 59.90 & 72.80 & 82.30 & 62.50 & 54.50 & 67.30 & 77.40 \\
RotateQVS  \cite{RotateQVS}             & 63.30  & 52.90 & 70.90 & 81.30 & 59.10 & 50.70 & 64.20 & 75.40 \\
TuckERTNT  \cite{TuckERTNT}               & 67.50  & 59.30 & 72.50 & 81.90  & --     & --     & --     & --     \\
TLT-KGE \cite{TLT-KGE}              & 60.90  & 60.90 & 74.10 & 83.50 & 63.40 & 55.10 & 68.40 & 78.60 \\
BoxTE \cite{BoxTE}                   & 66.70  & 58.20 & 71.90 & 82.00 & 61.30 & 52.80 & 66.40 & 76.30 \\
TKGC-AGP   \cite{TKGC-AGP}                & 53.20  & 39.80 & 62.10 & 79.70 & 56.10 & 45.80 & 63.10 & 73.80 \\
\hline
\end{tabular}
}
\vspace{-0.3 cm}
\label{TIPR}
\end{table}
\begin{table*}[!t]
\fontsize{40}{55}\selectfont 
    \caption{Performance comparison (in percentage) of temporal KGR models on GDELT, ICEWS14, ICEWS05-15, ICEWS18, WIKI, and YAGO for extrapolation scenario. Best results are marked as boldfaced. "H" is short for "Hits".}
  \vspace{-0.2cm}
    \resizebox{\linewidth}{!}{
\begin{tabular}{ccccccccccccccccccccccccc}
\hline
& \multicolumn{4}{c}{\textbf{GDELT}} & \multicolumn{4}{c}{\textbf{ICEWS14}} & \multicolumn{4}{c}{\textbf{ICEWS05-15}} & \multicolumn{4}{c}{\textbf{ICEWS18}} & \multicolumn{4}{c}{\textbf{WIKI}} & \multicolumn{4}{c}{\textbf{YAGO}} \\ \cline{2-25} 
\multirow{-2}{*}{\textbf{Model}} & \textbf{MRR}       & \textbf{H@1}     & \textbf{H@3}     & \textbf{H@10}    & \textbf{MRR}       & \textbf{H@1}     & \textbf{H@3}     & \textbf{H@10}   &\textbf{MRR}       & \textbf{H@1}     & \textbf{H@3}     & \textbf{H@10}     & \textbf{MRR}       & \textbf{H@1}     & \textbf{H@3}     & \textbf{H@10}    & \textbf{MRR}       & \textbf{H@1}     & \textbf{H@3}     & \textbf{H@10}   & \textbf{MRR}       & \textbf{H@1}     & \textbf{H@3}     & \textbf{H@10}    \\ \hline
RGCRN \cite{RGCRN}                   & 19.37     & 12.24      & 20.57      & 33.32      & 38.48      & 28.52      & 42.85      & 58.10       & 44.56      & 34.16       & 50.06       & 64.51        & 28.02      & 18.62      & 31.59      & 46.44       & 65.79     & 61.66      & 68.17     & 72.99      & 65.76     & 62.25      & 67.56     & 71.69      \\
RE-NET \cite{RE-NET}                 & 19.55     & 12.38      & 20.80      & 34.00      & 39.86      & 30.11      & 44.02      & 58.21       & 43.67      & 33.55       & 48.83       & 62.72        & 29.78      & 19.73      & 32.55      & 48.46       & 58.32     & 50.01      & 61.23     & 73.57      & 66.93     & 58.59      & 71.48     & 86.84      \\
CyGNet \cite{CyGNet}                   & 20.22     & 12.35      & 21.66      & 35.82      & 37.65      & 27.43      & 42.63      & 57.90       & 40.42      & 29.44       & 46.06       & 61.60        & 27.12      & 17.21      & 30.97      & 46.85       & 58.78     & 47.89      & 66.44     & 78.70      & 68.98     & 58.97      & 76.80     & 86.98      \\
TANGO     \cite{TANGO}                & 19.66     & 12.50      & 20.93      & 33.55      &  --          &  --           &    --         &    --            & 42.86      & 32.72       & 48.14       & 62.34        & 28.97      & 19.51      & 32.61      & 47.51       & 53.04     & 51.52      & 53.84     & 55.46      & 63.34     & 60.04      & 65.19     & 68.79      \\
xERTE  \cite{xERTE}                   & 19.45     & 11.92      & 20.84      & 34.18      & 40.79      & 32.70      & 45.67      & 57.30       & 46.62      & 37.84       & 52.31       & 63.92        & 29.31      & 21.03      & 33.51      & 46.48       & 73.60     & 69.05      & 78.03     & 79.73      & 84.19     & 80.09      & 88.02     & 89.78      \\
RE-GCN \cite{RE-GCN}                 & 19.69     & 12.46      & 20.93      & 33.81      & 42.00      & 31.63      & 47.20      & 61.65       & 48.03      & 37.33       & 53.90       & 68.51        & 32.62      & 22.39      & 36.79      & 52.68       & 78.53     & 74.50      & 81.59     & 84.70      & 82.30     & 78.83      & 84.27     & 88.58      \\
TLogic \cite{Tlogic}                 &  --          &  --           &    --         &    --           & 41.80      & 31.93      & 47.23      & 60.53       & 45.99      & 34.49       & 52.89       & 67.39        & 28.41      & 18.74      & 32.71      & 47.97        &  --          &  --           &    --         &    --           &  --          &  --           &    --         &    --          \\
CEN   \cite{CEN}                  &  --          &  --           &    --         &    --           & 41.64      & 31.22      & 46.55      & 61.59       & 49.57      & 37.86       & 56.42       & 71.32        & 29.70      & 19.38      & 33.91      & 49.90       & 63.39     & --          & 71.68     & 83.16      & 51.98     & --          & 58.96     & 70.61      \\
TITer \cite{TimeTraveler}                  & 18.19     & 11.52      & 19.20      & 31.00      & 41.73      & 32.74      & 46.46      & 58.44       & 47.60      & 38.29       & 52.74       & 64.86        & 29.98      & 22.05      & 33.46      & 44.83       & 73.91     & 71.70      & 75.41     & 76.96      & 87.47     & 80.09      & 89.96     & 90.27      \\
HisMatch \cite{HiSMatch}               & 22.01     & \textbf{14.45}      & 23.80      & 36.61      & \textbf{46.42}      & \textbf{35.91}      & \textbf{51.63}      & 66.84       & \textbf{52.85}      & \textbf{42.01}       & 59.05       & 73.28        & 33.99      & 23.91      & 37.90      & 53.94       & 78.07     & 73.89      & 81.32     & 84.65      &  --          &  --           &    --         &    --          \\
EvoKG     \cite{EvoKG}              & 19.28     & --          & 20.55      & 34.44      & 27.18      & --          & 30.84      & 47.67        &  --          &  --           &    --         &    --           & 29.28      & --          & 33.94      & 50.09       & 68.03     & --          & 79.60     & 85.91      & 68.59     & -          & 81.13     & 92.73      \\
TiRGN   \cite{TiRGN}                & 21.67     & 13.63      & 23.27      & 37.60      & 43.81      & 33.49      & 48.90      & 63.50       & 49.84      & 39.07       & 55.75       & 70.11        & 33.58      & 23.10      & 37.90      & 54.20       & 80.05     & 75.15      & 84.35     & 87.56      & 87.95     & 84.34      & 91.37     & 92.92      \\
RPC    \cite{RPC}                 & \textbf{22.41}     & 14.42      & \textbf{24.36}      & \textbf{38.33}      & 44.55      & 34.87      & 49.90      & 65.08       & 51.14      & 39.47       & 57.11       & 71.75        & \textbf{34.91}      & \textbf{24.34}     & 38.74      & 55.89       & \textbf{81.18}     & \textbf{76.28}      & \textbf{85.43}     & \textbf{88.71}      & \textbf{88.87}    & \textbf{85.10}      & \textbf{92.57}     & \textbf{94.04}      \\ 
CluSTeR  \cite{CluSTeR}                &  --          &  --           &    --         &    --             & 46.00      & 33.80      & -          & \textbf{71.20}      & 44.60      & 34.90       & --           & 63.00        & 32.30      & 20.60      & --          & 55.90       &  --          &  --           &    --         &    --            &  --          &  --           &    --         &    --          \\
RETIA  \cite{RETIA}                 & --        & --          & --          & --           & 45.29      & 34.60      & 50.88      & 66.06       & 52.17      & 40.21       & \textbf{59.42}       & \textbf{73.98}        & 34.16      & 22.97      & \textbf{39.27}     & \textbf{55.96}       & 67.58     & --          & 78.42     & 88.06      & 70.11     & --          & 78.30     & 84.77      \\
\hline
\end{tabular}
}
\vspace{-0.2 cm}
\label{TEPR}
\end{table*}

\paragraph{LSTM Enhanced Models}
Long Short-Term Memory (LSTM) network is also widely used to mine temporal features in temporal KGR models. For instance, TTransE \cite{TTransE} extends TransE by adding the temporal constraints and encodes time information as translations similar to relationships with an RNN so that these translations move the header representation in the embedded space. TA-TransE and TA-DistMult \cite{TA-DISTMULT} are also two respectively extended versions of TransE and DistMult that incorporate the temporal embeddings. Furthermore, EvolveGCN \cite{EvolveGCN} adopts the graph convolutional networks (GCNs) to model the graph structure in each static snapshot and utilizes the LSTM model (also can utilize GRU model) to evolve the GCN parameters over time. CluSTeR \cite{CluSTeR} adopts reinforcement learning to discover evolutional patterns with both LSTM and GRU models in Temporal KGs over time. DacKGR \cite{DacKGR} performs multi-hop path-based reasoning on sparse temporal KGs by using time information for dynamic prediction. To capture the timespan information and guide the model learning, TimeTraveler \cite{TimeTraveler} proposes a novel relative time encoding module and a time-shaped reward module based on Dirichlet distribution. TPath \cite{Tpath} also introduces LSTM model to mine the current environment information and then generates relation embeddings and temporal embeddings to the environment through activation functions. ExKGR \cite{ExKGR} introduces LSTM for reasoning in temporal KGs and provides the reasoning paths.

\paragraph{GRU Enhanced Models}
GRU-based models have got a lot of attention these years. More recently, TeMP \cite{TeMP2020} is proposed, which leverages message-passing graph neural networks (MPNNs) to learn structure-based entity representations at each timestamp, and then combines representations from all timestamps using an encoder. RE-GCN \cite{RE-GCN} focuses on the evolutional dynamics in temporal KGs and generates entity embeddings by modeling the KG sequence of a fixed length at the latest a few timestamps. TPmod \cite{TPmod} aggregates the attributes of entities and relations and learns dynamic weights to different events. HIP network \cite{HIP} passes information from temporal, structural, and repetitive perspectives, which are used to mine the graph's dynamic evolution, the interactions of events at the same time step, and the known events respectively. TRHyTE \cite{TRHyTE} uses GRU first to transform entities into latent space and then encode facts into temporal-relational hyper-planes for time relation-aware representation generation. TiRGN \cite{TiRGN} uses two encoders to mine the information at both local and global levels. HiSMatch \cite{HiSMatch} proposes different encoders to mine the semantic information of the historical query structures and candidate entities, respectively.

\subsubsection{RNN-agnostic Model}
RNN-agnostic models extend the original static KGR models by incorporating temporal information without using RNN frameworks. According to how the time information guides the models, they can be roughly divided into two types, \textit{i.e.,} time-vector guided and time-operation guided models.

\paragraph{Time-Vector Guided Model}
Time-vector guided models directly generate the additional temporal embedding $\textbf{t}$ as this additional temporal information and fuse them with fact embeddings.

TComplEx and TNTComplEx \cite{TNTComplEx} both come from ComplEx, where the fourth-order tensor space with additional consideration of time information is modeled by them. In the process of constructing subgraphs, the time embedding is used to calculate weighted probabilities in xERTE \cite{xERTE}. Then, T-GAP \cite{T-GAP} encodes the query-specific structure patterns of Temporal KG and performs path-based reasoning based on it. CyGNet \cite{CyGNet} attempts to solve the entity prediction task by encoding the historical facts related to the subject entity in each query and the time-indexing vector is generated. ChronoR \cite{ChronoR} builds on the basis of RotatE, which connects relation and time embeddings to obtain the overall rotation embedding applied to the final entity embedding. Furthermore, DBKGE \cite{DBKGE} proposed an online inference algorithm that smoothed the representation vector of nodes over time. BoxTE \cite{BoxTE} introduces a novel box representation method for temporal KGR based on the static KGR method BoxE. TuckERTNT \cite{TuckERTNT} proposes a novel tensor decomposition model for Temporal KGs inspired by the Tucker decomposition of a 4-order tensor with the extra time embedding. TempoQR \cite{TempoQR} generates question-specific time vectors and exploits these vectors to aggregate specific entities and their timestamps. TLT-KGE \cite{TLT-KGE} captures semantic and time information as different axes of complex space. RotateQVS \cite{RotateQVS} aims to consider the time information changes with rotation operation in the latent space.

\paragraph{Time-Operation Guided Model}
Time-operation guided models leverage some specific operations, such as encoding facts into designed time-specific hyper-planes and generating time-related rewards, to use the temporal information instead of directly fusion based on generating the temporal embeddings $\textbf{t}$.

ChronoTranslate \cite{ChronoTranslate} learns a universal representation of entities and time-specific representations of the Temporal KGs, respectively. HyTE \cite{HyTE} represents each timestamp as a learnable hyper-plane in the embedding space, then projects entity and relation embeddings into the hyper-plane and utilizes the TransE scoring function on the projections. As a model for graph learning and KGR, DyRep \cite{DyRep} captures the interleaved dynamics within history, which is further parameterized by a temporal-attentive representation network. Inspired by RotatE, TeRo \cite{TeRo} introduces a novel temporal guided rotation operation between head and tail entities to evaluate the given fact's semantic scores. Diachronic embeddings \cite{Diachronicembeddings} map entity and relation embeddings, paired with temporal information, into a KGR model space, thus defining a framework yielding specific models such as DE-TransE and DE-SimplE. Due to the uncertainty of temporal information in the graph evolution over time, ATiSE \cite{ATiSE} maps the entity and relation embeddings of Temporal KGs into the Gaussian spaces according to the time stamps. TDGNN \cite{TDGNN} introduces a novel temporal aggregator to combine the neighborhood features and the temporal information from edges to calculate the final representations. To mine the dynamic graph evolution of temporal KGs, DyERNIE \cite{DyERNIE} defines the velocity vector in the tangent space over time and encourages entity embeddings to evolve according to it. TPRec \cite{TPRec} is an interest recommendation method that presents an efficient time-aware interaction relation extraction component to construct a collaborative KG with time-aware interactions and also utilizes a time-aware path module for reasoning. TeLM \cite{TeLM} leverages a linear temporal regularizer and multi-vector encoders to realize the 4th-order tensor factorization for reasoning. RTFE \cite{RTFE} treats the Temporal KGs as a Markov chain, which transitions from the previous state to the next state, and then recursively tracks the state transition of Temporal KG by passing updated parameters/features between timestamps. Besides, TIE \cite{TIE} combines the experience replay and time regularization into KGR to learn the time-aware incremental embedding. A length-aware CNN is leveraged in CEN \cite{CEN} to handle historical facts via an easy-to-difficult curriculum learning strategy over time. DKGE \cite{DKGE} introduces two different representations for each entity and each relationship (including temporal information).  TLogic \cite{Tlogic} performs the temporal random walks and extracts the temporal logical rules based on them, leading to better explainability. TKGC-AGP \cite{TKGC-AGP} leverages the approximations of multivariate Gaussian processes (MGPs) for fact encoding. Besides, FILT \cite{FILT} makes use of the meta-learning framework for inferring the facts with unseen entities in temporal KGR. After that, rGalT \cite{rGalT} first design the attention mechanism in both intra-graph and inter-graph levels to leverage the historical semantics. Similarly to it, DA-Net \cite{DA-Net} also tries to learn attention weights on repetitive facts at different historical timestamps. MetaTKGR \cite{MetaTKGR} dynamically adjusts the strategies of sampling and aggregating neighbors from recent facts for new entities through temporally supervised signals on future facts as instant feedback. Moreover, CENET \cite{CENET} learns both the historical and non-historical dependency for inferring the most potential facts.

\vspace{-0.15 cm}
\subsection{Review on Reasoning Scenarios}
Based on the observation, there are \textbf{34} interpolation models and \textbf{24} extrapolation models (See Table \ref{Sum_Tem}). Among them, 42.57\% of the extrapolation models belong to time-operation models, and 47.62\% belongs to RNN-based models. In particular, we observe that among the RNN-based models, the ratio of interpolation models and extrapolation models is really close (10:8). Meanwhile, similar observations can also be found among Time-Operation models (12:9). It reveals that RNN-based models and time-Operation models have shown good compatibility with both scenarios. Also, most of the recent research lies in this type. Compared to them, time-vector models can only perform better in interpolation scenarios since they cannot sufficiently the temporal information in such a primitive manner. Moreover, we present a fair performance comparison of the typical SOTA temporal KGR models (See Table \ref{TIPR} and Table \ref{TEPR}). Similar conclusions can be drawn based on the results, which support the analysis above.

\subsection{Observation and Discussion}
Based on the above reviews, we can further get the following observations, which may indicate the scope for different temporal KGR models and reveal the future trend in temporal KGR. (1) RNN-agnostic models, \textit{i.e.,} time-vector guided models and time-operation guided models, generally treat the temporal information as additional attributes and integrate them into the previous static KGR models with different techniques. Such a manner is more flexible compared to RNN-based models. (2) Time-vector guided models encode the time information as an additional time vector $\textbf{t}$. Although these models are simple, their performances mostly rely on whether the time encoder and embedding fusion module are suitable. Unlike these models, time-operation guided models design specific time operations, which are task-specific. (3) RNN-based models can generally model the time information better than other models and can be more easily adopted to extrapolation scenarios. (4) The extrapolation reasoning is still at an early stage, occupying only around $30\%$ of the temporal KGR models, which leaves space for further exploration. As a brief conclusion, we would encourage to study more on RNN-based models, which are considered as the most promising model type for temporal KGR.

\vspace{0.1 cm}
\section{Multi-Modal KGR Model}
\begin{figure}[!t]
\setlength{\abovecaptionskip}{-0.02cm}
\centering
\includegraphics[width=0.46\textwidth]{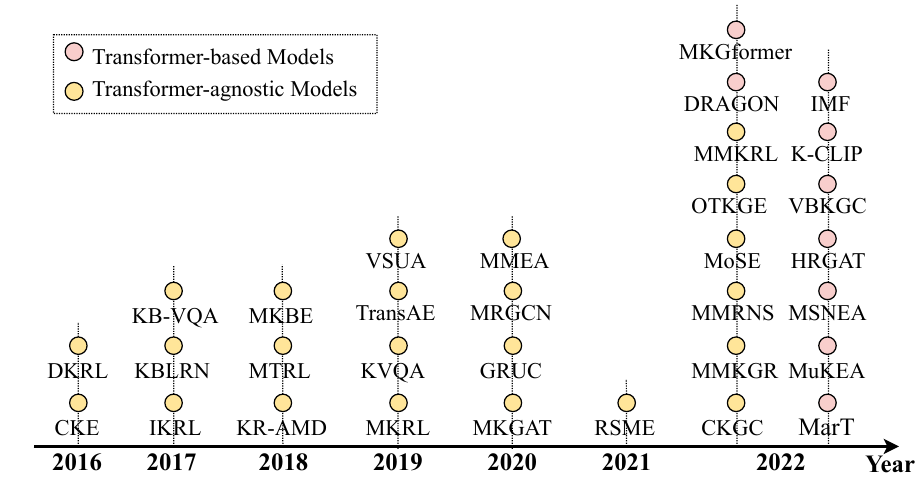}
\vspace{0.1 cm}
\caption{Timeline of the multi-modal KGR models.}
\vspace{-0.3 cm}
\label{MKGTim}  
\end{figure}

We systematically introduce \textbf{32} multi-modal KGR models according to techniques (See Figure \ref{MKGTim}).

\vspace{-0.1 cm}
\subsection{Review on Reasoning Techniques}
Directly applying static KGR models to multi-modal scenarios generally results in sub-optimal performance because of the lack of fusion modules for extra multi-modal information, \textit{e.g.,} texts, images, etc. Based on techniques to fuse such multi-modal information, we roughly divided the multi-modal KGR models into two types, \textit{i.e.,} transformer-based and transformer-agnostic models.

\subsubsection{Transformer-based Models}
Transformer-based models are usually adopted as the unified paradigm for multi-modal problems due to their promising capacity when scaling to different modalities. 

Although some general multi-modal pretrained transformer models, such as VisualBERT \cite{visualbert}, ViLBERT \cite{VL-Bert}, can also be adopted for multi-modal KGR. Due to the variance between the multi-modal KGs and other multi-modal data, directly applying the above general MPT models to multi-modal knowledge graph reasoning (MKGR) may not lead to good reasoning performance. Inspired by it, researchers have attempted to develop transformer-based multi-modal KGR models these two years. VBKGC \cite{VBKGC} leverages the pretrained transformer to encode multi-modal features and designs a multi-modal scoring function for optimization. Then, Knowledge-CLIP \cite{Knowledge-CLIP} leverages the CLIP \cite{CLIP} model for a better pre-trained model considering the semantic connections between multi-modal concepts. Meanwhile, MarT \cite{MARS} first proposes a model-agnostic reasoning framework with a transformer for analogical reasoning. Additionally, a hybrid transformer with multi-level fusion is designed in MKGformer \cite{MKGformer}, which unifies learning paradigms for different downstream tasks in a uniform framework. Later on, HRGAT \cite{HRGAT} constructs a hyper-node graph to aggregate the multi-modal features generated by transformers. Besides, MSNEA \cite{MSNEA} and IMF \cite{IMF} makes use of contrastive learning for multi-modal alignment. Moreover, MuKEA \cite{MuKEA} integrates KGR in vision understanding and reasoning, and DRAGON \cite{DRAGON} also provides a method for pre-training in a self-supervised manner for text and KG. However, transformer-based multi-modal KGR is still at an early stage.

\subsubsection{Transformer-agnostic Models}
Most multi-modal KGR models generate and fuse features without using transformer frameworks but design different mechanisms to encode the extra modal information by extending the original unimodal KGR models, such as TransE \cite{TransE}. These models are named transformer-agnostic models.

CKE \cite{CKE} is the first model to perform reasoning and collaborative filtering jointly, which enables it to generate representations and capture the implicit rules in KGs simultaneously. Then, DKRL \cite{DKRL1} takes advantage of entity descriptions in KGs with language neural networks and gets more expressive semantics for reasoning. Inspired by it, IKRL \cite{IKRL} first designs an attention-based neural network to consider visual information in entity images. Such attention mechanism is also leveraged by TransAE \cite{TransAE}. KBLRN \cite{KBLRN} first proposes an end-to-end reasoning framework, which combines neural network techniques with expert models for latent, relational, and numerical features. Afterward, KR-AMD \cite{KR-AMD} and MKRL \cite{MKRL} leverage textual data as part of auxiliary data to improve reasoning performance. Besides, inspired by the translation-based static KGR models, MTRL \cite{MM-TransE} is a translation-based model with three energy functions corresponding to visual, linguistic, and structural information. Moreover, MKBE \cite{MKBE} and MRCGN \cite{MRCGN} integrate different neural encoders and decoders with relational models for embedding learning and multi-modal data for reasoning. MMKGR \cite{MMKGR} first investigates the problem of how to effectively leverage multi-modal auxiliary features to conduct multi-hop reasoning in the KG area with a unified gate-attention network. MKGAT \cite{MKGAT} better enhances recommendation systems with a multi-modal graph attention technique to conduct information propagation over multi-modal KGs. KB-VQA \cite{KB-VQA} and VSUA \cite{VSUA} performs reasoning on the image and external knowledge, which provides an intuitive way to explain the generated answers. Similar to it, KVQA \cite{KVQA} integrates commonsense knowledge with images for reasoning. MMEA \cite{MMEA} designs a joint loss for multi-modal alignment. Moreover, MoSE \cite{MoSE} exploits three ensemble inference techniques to combine the modality-split predictions by assessing modality importance. Recently, RSME \cite{RSME} designed a forget gate with an MRP metric to select valuable images for multi-modal KGR, which tries to avoid the influence caused by the noise from irrelevant images corresponding to entities. HMEA \cite{HMEA} projects multi-modal features into a hyperbolic space via GCN. While, OTKGE \cite{Otkge} models the multi-modal fusion procedure as a transport plan moving different modal embeddings to a unified space by minimizing the Wasserstein distance. Besides, MM-RNS \cite{MM-RNS} and CKGC \cite{CKGC} leverage contrastive learning strategies. MMKRL \cite{MMKRL} leverages reinforcement learning for multi-modal KGR.

\begin{table}[t]
\caption{Performance comparison (in percentage) of multi-modal KGR models on FB15K-237-IMG and WN18-IMG. The best results are in boldface. "H" is short for "Hits".}
\vspace{-0.2 cm}
\fontsize{7}{8.5}\selectfont 
\resizebox{0.98\linewidth}{!}{
\begin{tabular}{ccccccccc}
\hline
\multirow{2}{*}{\textbf{Model}}   & \multicolumn{4}{c}{\textbf{FB15k-237-IMG}} & \multicolumn{4}{c}{\textbf{WN18-IMG}} \\ \cline{2-5} \cline{6-9} 
 & \textbf{MR} & \textbf{H@1}    & \textbf{H@10}      & \textbf{MR} & \textbf{H@1} & \textbf{H@10}  \\ \hline
IKRL  \cite{IKRL}           & 298 & 19.4   & 45.8     & 596 & 12.7  & 92.8   \\
TransAE  \cite{TransAE}               & 431 & 19.9    & 46.3    & 352& 32.3  & 93.4  \\
MTRL  \cite{MM-TransE}             & 187 & {22.9}     & {49.4}      & --  & --  & -- \\
MKBE  \cite{MKBE}             & 158 & {25.8}     & {53.2}     & --  & --  & -- \\
RSME  \cite{RSME}  & 417 & 24.2   & 46.7     &223& {94.3}  & 95.7    \\  
MoSE   \cite{MoSE}            & \bf{117} & {28.1}     & {56.5}     & \bf{7} & \bf{94.8}  & \bf{97.4} \\
KBLRN  \cite{KBLRN}            & 209 & {21.9}     & {49.3}      & --  & --  & -- \\
VisualBERT  \cite{visualbert}          & 592 & 21.7    & 43.9     & 122 & 17.9  & 65.4   \\
ViLBERT \cite{VL-Bert}               & 483 & 23.3     & 45.7    & 131  & 22.3 & 76.1  \\
VBKGC  \cite{VBKGC}  & - & 21.3   & 47.8     & --  & --  & -- \\
HRGAT  \cite{HRGAT}            & 156 & {27.1}     & {54.2}     & --  & --  & -- \\
MKGformer  \cite{MKGformer}            & 252 & {24.3}     & {49.9}     & {25}  & {93.5}  & {97.0}  \\
IMF \cite{IMF}            & 134 & \bf{28.7}     & \bf{59.3}     & --  & --  & -- \\\hline
\end{tabular}
}
\label{Mper} 
\end{table}
\begin{figure}[t]
\centering
\includegraphics[width=0.48\textwidth]{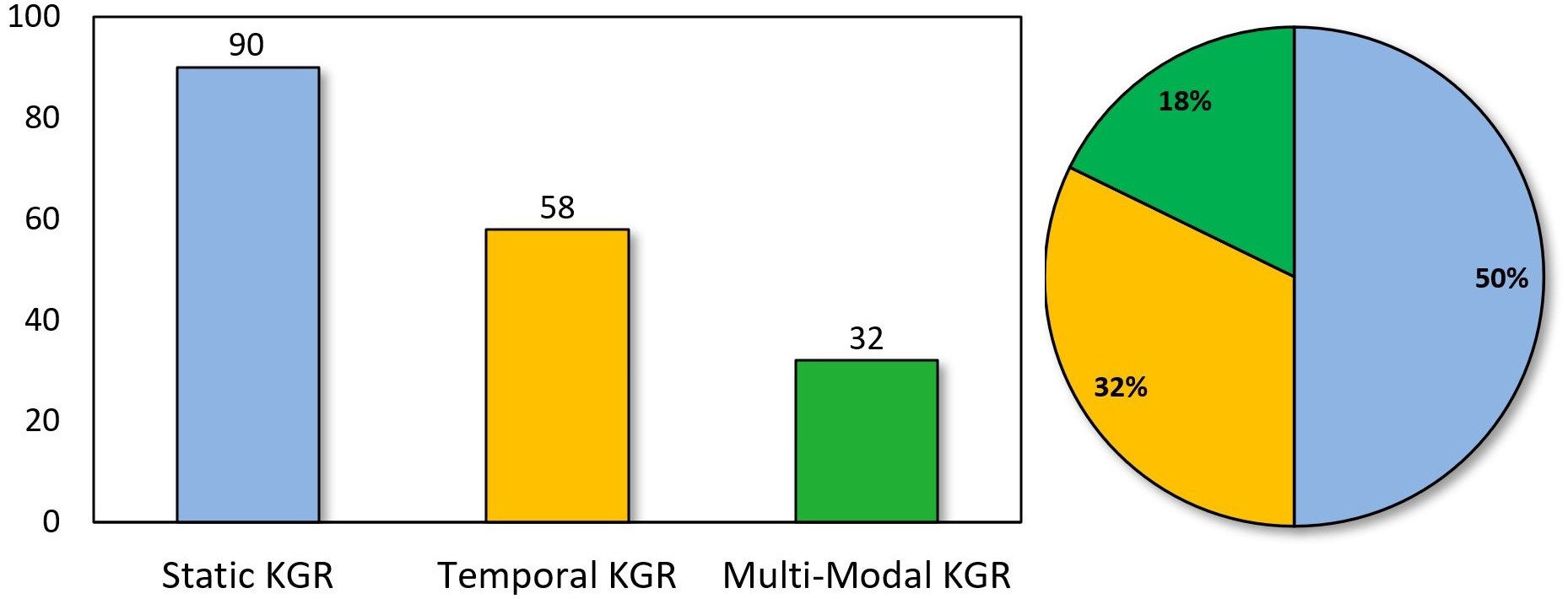}
\caption{Statistic comparison of models over various KG types.}
\label{StatGraph} 
\vspace{-0.2 cm}
\end{figure}

\subsection{Observation and Discussion}
Based on the above reviews and performance comparison (See Table \ref{Mper}), we can further get the following observations. (1) Initially, most multi-modal KGR models are developed based on embedding-based static KGR models instead of path-based or rule-based ones. It is mainly because most existing multi-modal KGR models leverage the extra multi-modal information via feature fusion in latent space. Specifically, different encoders are designed for features in different modalities. (2) Recently, researchers have tended to study uniform learning frameworks, such as pre-trained transformer-based models, for multi-modal features, especially after Large Language Models (LLM) flourished. These models meet the requirements of artificial general intelligence, which are more practical and scalable in this era. (3) Compared to the other two types of KGR, the research on multi-modal KGR is still at an early stage, \textit{i.e.,} only 18\% models are for multi-modal scenarios. In conclusion, there are lots of spaces for deep exploration, further described in Sec. 7.4 as a challenge and opportunity.

\section{Datasets}
We comprehensively summarize typical KGR datasets, especially for temporal and multi-modal KGs, and provide their description and statistic as follows. Besides, the datasets are collected in our GitHub repository to better convenience the community.

\vspace{-0.15 cm}
\subsection{Static KGR Datasets}
Typical static KGR datasets, \textit{i.e.,} \textbf{38} transductive datasets, and \textbf{15} inductive datasets are summarized. The statistics are presented in Table \ref{STR_dataset} and Table \ref{SIR_dataset}, and the descriptions are listed below.
\begin{table}[t]
\fontsize{21.5}{25}\selectfont 
\caption{Typical benchmark datasets for static transductive knowledge graph reasoning.}
\vspace{-0.2 cm}
\resizebox{\linewidth}{!}{
\begin{tabular}{cccccc}
\hline \textbf{Dataset} & \textbf{\# Ent.} & \textbf{\# Rel.} & \textbf{\# Train Facts} & \textbf{\# Val. Facts} & \textbf{\# Test Facts}\\ 
\hline
ATOMIC \cite{Atomic}         & 304,388              & 9                    & 610,536                    & 87,700                    & 87,701                    \\ 
Countries \cite{Countries}       & 271                 & 2                    & 1,110                      & 24                       & 24                       \\ 
CoDEX-S  \cite{CoDEx}        & 2,034                & 42                   & 32,888                     & 3,654                     & 3656                     \\ 
CoDEX-M   \cite{CoDEx}          & 17,050               & 51                   & 185,584                    & 20620                    & 20622                    \\ 
CoDEX-L  \cite{CoDEx}           & 77,951               & 69                   & 551,193                    & 30,622                    & 30622                    \\ 
ConceptNet  \cite{Conceptnet1}        & 28,370,083            & 50                   & 27,259,933                  & 3,407,492                  & 3,407,492                  \\ 
ConceptNet100K  \cite{BiQUE}   & 78,334               & 34                   & 100,000                    & 1,200                     & 1,200                     \\ 
DBpedia50   \cite{DBpedia50}     & 49,900               & 654                  & 32,388                     & 399                      & 10,969                    \\ 
DBpedia500   \cite{DBpedia50}     & 517,475              & 654                  & 3,102,677                   & 10,000                    & 1,155,937                  \\ 
DB100K   \cite{DB100K}         & 99,604               & 470                  & 597,482                    & 49,997                    & 50,000                    \\ 
FAMILY    \cite{family}         & 3,007                & 12                   & 23,483                     & 2,038                     & 2,835                     \\ 
FB13    \cite{FB13}         & 75,043               & 13                   & 316,232                    & 11,816                    & 47,464                    \\ 
FB122   \cite{FB122}         & 9,738                & 122                  & 91,638                     & 9,595                     & 11,243                    \\ 
FB15k   \cite{FB15k}         & 14,951               & 1,345                 & 483,142                    & 50,000                    & 59,071                    \\
FB20k \cite{DBpedia50}           & 19,923               & 1,345                 & 472,860                    & 48,991                    & 90,149                    \\
FB24k \cite{FB24k}           & 23,634
           & 673                 & 402,493                   & -                   & 21,067                   \\
FB5M  \cite{TransH}      &   5,385,322               & 1,192                  &  19,193,556                   & 50,000                    &  59,071                 \\
FB15k-237  \cite{FB15k-237}      & 14,505               & 237                  & 272,115                    & 17,535                    & 20,466                    \\
FB60k-NYT10  \cite{FB60k-NYT10}     & 69,514               & 1,327                 & 268,280                    & 8,765                     & 8,918                     \\ 
Hetionet   \cite{Hetionet}      & 45,158               & 24                   & 1,800,157                   & 225,020                   & 225,020                   \\
Kinship   \cite{family}       & 104                 & 25                   & 8,544                      & 1,068                     & 1,074                     \\ 
Location  \cite{sports}       & 445                 & 5                    & 384                       & 65                       & 65                       \\
Nation     \cite{nation}      & 14                  & 55                   & 1,592                      & 199                      & 201                      \\
NELL23k   \cite{NELL23k}       & 22,925               & 200                  & 25,445                     & 4,961                     & 4,952                     \\
NELL-995   \cite{NELL-995}      & 75,492               & 200                  & 126,176                    & 5,000                     & 5,000                     \\ 
OpenBioLink  \cite{OpenBioLink}    & 180,992              & 28                   & 4,192,002                   & 188,394                   & 183,011                   \\ 
Sport   \cite{sports}         & 1,039                & 4                    & 1,349                      & 358                      & 358                      \\
Toy  \cite{libkge}            & 280                 & 112                  & 4,565                      & 109                      & 152                      \\ 
UMLS  \cite{UMLS-1}           & 135                 & 46                   & 5,216                      & 652                      & 661                      \\
UMLS-PubMed  \cite{FB60k-NYT10}    & 59,226               & 443                  & 2,030,841                   & 8,756                     & 8,689                     \\
WD-singer   \cite{NELL23k}     & 10,282               & 135                  & 16,142                     & 2,163                     & 2,203                     \\
WN11     \cite{FB13}        & 38,588               & 11                   & 110,361                    & 5,212                     & 21,035                    \\ 
WN18     \cite{WN18}        & 40,943               & 18                   & 141,442                    & 5,000                     & 5,000                     \\ 
WN18RR   \cite{FB15k-237}        & 40,559               & 11                   & 86,835                     & 2,924                     & 2,824                     \\
wikidata5m  \cite{wikidata5m}     & 4,594,485             & 822                  & 20,614,279                  & 5,163                     & 5,163                     \\
YAGO3-10    \cite{yago3}     & 123,143              & 37                   & 1,079,040                   & 4,978                     & 4,982                     \\ 
YAGO37 \cite{yogo37}     & 123,189              & 37                   & 420,623                    & 50,000                    & 50,000                    \\ 
M-/YAGO39k  \cite{yogo39k}     & 85,484               & 39                   & 354,997                    & 9,341                     & 9,364                     \\
\hline
\end{tabular}
}
\label{STR_dataset}
\vspace{-0.1 cm}
\end{table}

\begin{table}[t]
\fontsize{24}{26.5}\selectfont 
\caption{Typical benchmark datasets for static inductive knowledge graph reasoning.}
\vspace{-0.2 cm}
\resizebox{\linewidth}{!}{
\begin{tabular}{ccccccc}
\hline
\multicolumn{2}{c}{\textbf{Dataset}}                  & \textbf{\# Ent.} & \textbf{\# Rel.} & \textbf{\# Train Facts} & \textbf{\# Val. Facts} & \textbf{\# Test Facts} \\ 
\hline
\multirow{2}{*}{WN18RR v1 \cite{GraIL}}   & train-graph      & 2,746       & 9           & 5,410             & 626             & 638                       \\
& ind-test-graph   & 922        & 9           & 1,618             & 181             & 184                       \\
\multirow{2}{*}{WN18RR v2 \cite{GraIL}}   & train-graph      & 6,954       & 10          & 15,262            & 1,837            & 1,868                      \\
& ind-test-graph   & 2,923       & 10          & 4,011             & 407             & 437                       \\
\multirow{2}{*}{WN18RR v3 \cite{GraIL}}   & train-graph      & 12,078      & 11          & 25,901            & 3,097            & 3,152                      \\
& ind-test-graph   & 5,084       & 11          & 6,327             & 534             & 601                       \\
\multirow{2}{*}{WN18RR v4 \cite{GraIL}}   & train-graph      & 3,861       & 9           & 7,940             & 934             & 968                       \\
& ind-test-graph   & 7,208       & 9           & 12,334            & 1,394            & 1,429                      \\
\multirow{2}{*}{FB15k237 v1 \cite{GraIL}} & train-graph      & 2,000       & 183         & 4,245             & 485             & 492                       \\
& ind-test-graph   & 1,500       & 146         & 1,993             & 202             & 201                       \\
\multirow{2}{*}{FB15k237 v2 \cite{GraIL}} & train-graph      & 3,000       & 203         & 9,739             & 1,166            & 1,180                      \\
& ind-test-graph   & 2,000       & 176         & 4,145             & 469             & 478                       \\
\multirow{2}{*}{FB15k237 v3 \cite{GraIL}} & train-graph      & 4,000       & 218         & 17,986            & 2,194            & 2,214                      \\
& ind-test-graph   & 3,000       & 187         & 7,406             & 866             & 865                       \\
\multirow{2}{*}{FB15k237 v4 \cite{GraIL}} & train-graph      & 5,000       & 222         & 27,203            & 3,352            & 3,361                      \\
& ind-test-graph   & 3,500       & 204         & 11,714            & 1,416            & 1,424                      \\
\multirow{2}{*}{NELL995 v1 \cite{GraIL}}  & train-graph      & 10,915      & 14          & 4,687             & 414             & 435                       \\
& ind-test-graph   & 225        & 14          & 833              & 97              & 96                        \\
\multirow{2}{*}{NELL995 v2 \cite{GraIL}}  & train-graph      & 2,564       & 88          & 8,219             & 922             & 968                       \\
& ind-test-graph   & 4,937       & 79          & 4,586             & 455             & 476                       \\
\multirow{2}{*}{NELL995 v3 \cite{GraIL}}  & train-graph      & 4647       & 142         & 16,393            & 1,851            & 1,873                      \\
& ind-test-graph   & 4,921       & 122         & 8,048             & 811             & 809                       \\
\multirow{2}{*}{NELL995 v4 \cite{GraIL}}  & train-graph      & 2,092       & 77          & 7,546             & 876             & 867                       \\
& ind-test-graph   & 3,294       & 61          & 7,073             & 716             & 731                       \\
\multirow{6}{*}{WN-MBE \cite{MBE}}     & train-graph      & 19,361      & 11          & 35,426            & 8,858            & -                         \\
& ind-test-graph-1 & 3,723       & 11          & 5,678             & -               & 1,352                      \\
& ind-test-graph-2 & 4,122       & 11          & 6,730             & -               & 1,874                      \\
& ind-test-graph-3 & 4,300       & 11          & 7,545             & -               & 2,054                      \\
& ind-test-graph-4 & 4467       & 11          & 8,623             & -               & 2,493                      \\
& ind-test-graph-5 & 4,514       & 11          & 9,608             & -               & 2,762                      \\
\multirow{6}{*}{FB-MBE \cite{MBE}}     & train-graph      & 7,203       & 237         & 125,769           & 31,442           & -                         \\
& ind-test-graph-1 & 1,458       & 237         & 18,394            & -               & 9,240                      \\
& ind-test-graph-2 & 1,461       & 237         & 19,120            & -               & 9,669                      \\
& ind-test-graph-3 & 1,467       & 237         & 19,740            & -               & 9,887                      \\
& ind-test-graph-4 & 1,467       & 237         & 22,455            & -               & 11,127                     \\
& ind-test-graph-5 & 1,471       & 237         & 22,214            & -               & 11,059                     \\
\multirow{6}{*}{NELL-MBE \cite{MBE}}   & train-graph      & 33,348      & 200         & 88,814            & 22,203           & -                         \\
& ind-test-graph-1 & 34,488      & 3,200        & 34,496            & -               & 3,853                      \\
& ind-test-graph-2 & 36031      & 3,200        & 35,411            & -               & 31,059                     \\
& ind-test-graph-3 & 37,660      & 3,200        & 36,543            & -               & 31,277                     \\
& ind-test-graph-4 & 39,056      & 3,200        & 37,667            & -               & 31,427                     \\
& ind-test-graph-5 & 310,616     & 3,200        & 38,876            & -               & 31,595\\
\hline
\end{tabular}}
\label{SIR_dataset}
\vspace{-0.2 cm}
\end{table}

\begin{itemize}[leftmargin=0.5cm]
    \item \textbf{ATOMIC} \cite{Atomic} is an KGs for everyday commonsense reasoning. It is composed of the reactions, effects, and intents of human behaviors and descriptions of each entity.

    \item \textbf{Countries} \cite{Countries} consists of relations among countries based on public geographical data.

    \item \textbf{CoDEX} \cite{CoDEx} is a set of COmpletion Datasets EXtracted from Wikidata and Wikipedia, which contains three different sizes of sub-KGs, \textit{i.e.,} CoDEX-S, CoDEX-M, CoDEX-L.
   
    \item \textbf{Conceptnet} \cite{Conceptnet1} connects words and phrases with labeled edges to enhance AI APPs to understand word meanings better. \textbf{Conceptnet100K} \cite{BiQUE} contains 100k training triplets.

    \item \textbf{DBpedia} \cite{DBpediafull} consists of structured content from the information created in various Wikimedia projects. According to the entity set size, we can derive several subsets from it, \textit{i.e.,} \textbf{DBpedia50} \cite{DBpedia50}, \textbf{DBpedia500} \cite{DBpedia50} and \textbf{DB100K} \cite{DB100K}.

    \item \textbf{FAMILY} \cite{family} consists of relations among family members.

    \item \textbf{FreeBASE} \cite{Freebase} is a large knowledge base generated from multiple sources, such as Wikipedia, NNDB, Fashion Model Directory, etc. According to the entity set size, we can derive several subsets from it, including \textbf{FB13} \cite{FB13}, \textbf{FB122} \cite{FB122}, \textbf{FB15k} \cite{FB15k}, \textbf{FB20k} \cite{DBpedia50}, \textbf{FB24k} \cite{FB24k}, \textbf{FB5M} \cite{TransH}, \textbf{FB15k-237} \cite{FB15k-237}, \textbf{FB60k-NYT10} \cite{FB60k-NYT10}.
   
    \item \textbf{Hetionet} \cite{Hetionet} is a knowledge graph derived from biomedical studies based on public resources. It describes relations among compounds, diseases, genes, anatomies, pathways, biological processes, molecular functions, cellular components, pharmacologic classes, side effects, and symptoms.

    \item \textbf{Kinship} \cite{family} describe kinships in Alyawarra tribes \cite{Denham}. 

    \item \textbf{Nation} \cite{family} contains relations among nations \cite{nation}. 

    \item \textbf{NELL} \cite{NELL} is the knowledge base built based on Never-Ending Language Learner, which attempts to learn to read the web over time. According to the entity set size, we can derive several subsets from it, \textit{e.g.,} \textbf{Location} \cite{sports}, \textbf{sports} \cite{sports}, \textbf{NELL23k} \cite{NELL23k}, \textbf{NELL-995} \cite{NELL-995}.

    \item \textbf{OpenBioLink} \cite{OpenBioLink} is a large-scale, high-quality, and highly challenging biomedical KG. 

    \item \textbf{Toy} \cite{libkge} is a small KG used for testing and debugging.

    \item \textbf{UMLS} \cite{UMLS} is the KG of the Unified Medical Language System. By cooperating with the PubMed corpus, it is extended to \textbf{UMLS-PubMed} \cite{FB60k-NYT10}.

    \item \textbf{WordNet} \cite{WordNet} is a lexical database of semantic relations, \textit{e.g.,} synonyms, hyponyms, and meronyms, between words. According to the entity set size, we can derive several subsets from it, \textit{e.g.,} \textbf{WN11} \cite{FB13}, \textbf{WN18} \cite{WN18}, \textbf{WN18RR} \cite{FB15k-237}.

    \item \textbf{Wikidata} \cite{Wikidata} provides common sources for Wikipedia, where \textbf{WD-singer} \cite{NELL23k} and \textbf{wikidata5m} \cite{wikidata5m} are subsets.
    
    \item \textbf{YAGO} \cite{Yago}, as a lightweight and extensible ontology, is built from {Wikidata} and unified with {WordNet}. According to the sizes of relations, \textbf{YAGO3-10} \cite{yago3}, \textbf{YAGO37} \cite{yogo37} and \textbf{YAGO39k} \cite{yogo39k} can be derived.   
\end{itemize}

\subsection{Temporal KGR Datasets}
Eighteen Typical temporal KGR datasets are summarized. The statistic is presented in Table \ref{TR_dataset}, and descriptions are listed below.
\begin{table}[t]
\fontsize{30}{43}\selectfont 
\caption{Typical benchmark datasets for temporal knowledge graph reasoning.}
\vspace{-0.2 cm}
\resizebox{\linewidth}{!}{
\begin{tabular}{ccccccc}
\hline
\textbf{Dataset}                  & \textbf{\# Ent.} & \textbf{\# Rel.} &\textbf{\# Timestamps} & \textbf{\# Train Facts} & \textbf{\# Val. Facts} & \textbf{\# Test Facts} \\ 
\hline
DBpedia-3SP  \cite{DBpedia-3SP}    & 66,967      & 968         & 3            & 103,211           & 3,000            & -               \\
GDELT  \cite{TLT-KGE}           & 7,691       & 240         & 8,925         & 1,033,270          & 238,765          & 305,241          \\
GDELT-small \cite{TIE}      & 500        & 20          & 366          & 2,735,685          & 341,961          & 341,961          \\
GDELT-m10  \cite{GDELT-m10}       & 50         & 20          & 30           & 221,132           & 27,608           & 27,926           \\
IMDB-13-3SP \cite{TLT-KGE}      & 3,244,455    & 14          & 3            & 7,913,773          & 10,000           & -               \\
IMDB-30SP \cite{TLT-KGE}        & 243,148     & 14          & 30           & 621,096           & 3,000            & 3,000            \\
ICEWS05-15 \cite{ICEWS}       & 10,488      & 251         & 4,017         & 386,962           & 46,092           & 46275           \\
ICEWS11-14  \cite{ICEWS}      & 6,738       & 235         & 1,461         & 118,766           & 14,859           & 14,756           \\
ICEWS14   \cite{xERTE}        & 7,128       & 230         & 365          & 63,685            & 13,823           & 13,222           \\
ICEWS14-Plus \cite{GDELT-m10}     & 7,128       & 230         & 365          & 72,826            & 8,941            & 8,963            \\
ICEWS18   \cite{xERTE}        & 23,033      & 256         & 7,272         & 373,018           & 45,995           & 49,545           \\
YOGA11k/YOGA \cite{HyTE}     & 10,623      & 10          & 189          & 161,540           & 19,523           & 20,026           \\
YOGA-3SP  \cite{DBpedia-3SP}        & 27,009      & 37          & 3            & 124,757           & 3,000            & 3,000            \\
YOGA15k   \cite{ICEWS}        & 15,403      & 34          & 198          & 110,441           & 13,815           & 13,800           \\
YOGA1830  \cite{xERTE}        & 10,038      & 10          & 205          & 51,205            & 10,973           & 10,973           \\
WIKI/Wikidata12k \cite{HyTE} & 12,554      & 24          & 232          & 2,735,685          & 341,961          & 341,961          \\
Wikidata11k \cite{T-GAP}      & 11,134      & 95          & 328          & 242,844           & 28,748           & 14,283           \\
Wikidata-big  \cite{TempoQR}    & 125,726     & 203         & 1,700         & 323,635           & 5,000            & 5,000     \\
\hline
\end{tabular}}
\label{TR_dataset}
\vspace{-0.2 cm}
\end{table}
\begin{itemize}[leftmargin=0.5cm]
    \item \textbf{DBpedia-3SP} \cite{DBpedia-3SP} is extracted subsets from \textbf{DBpedia} in three different timestamps.

    \item \textbf{GDELT} \cite{TLT-KGE} is a dense KG derived from the Global Database of Events, Language, and Tone. \textbf{GDELT-m10} \cite{GDELT-m10} and \textbf{GDELT-small} \cite{TIE} are extracted from it.

    \item \textbf{IMDB} \cite{IMDB} is a KG consisting of the entities of movies, TV series, actors, and directors, which is also known as the Internet Movie Database. \textbf{IMDB-30SP} \cite{TLT-KGE} and \textbf{IMDB-13-3SP} \cite{TLT-KGE} are extracted from the dataset in different timestamps.

    \item \textbf{ICEWS} \cite{ICEWS-FULL}, short for Integrated Crisis Early Warning System, is a database that contains political events with specific timestamps. Some typical temporal KGs are created out of it, \textit{i.e.,} \textbf{ICEWS05-15} \cite{ICEWS}, \textbf{ICEWS11-14} \cite{ICEWS}, \textbf{ICEWS14} \cite{xERTE}, \textbf{ICEWS14-Plus} \cite{GDELT-m10}, \textbf{ICEWS18} \cite{xERTE}.

    \item \textbf{Wikidata} \cite{Wikidata} for temporal KGR contains extra time information than the static Wikidata dataset. \textbf{WIKI/Wikidata12k} \cite{HyTE}, \textbf{Wikidata11k} \cite{T-GAP} and \textbf{Wikidata-big} \cite{TempoQR} are generated based on different periods.

    \item \textbf{YAGO} \cite{Yago} for temporal KGR contains extra time information. \textbf{YOGA11k/YOGA} \cite{HyTE}, \textbf{YOGA15k} \cite{ICEWS}, \textbf{YOGA-3SP} \cite{DBpedia-3SP} and \textbf{YOGA1830} \cite{xERTE} are generated from it according to different periods.
\end{itemize}

\vspace{-0.2 cm}
\subsection{Multi-Modal KGR Datasets}
Eleven typical multi-modal KGR datasets are summarized. The statistic is presented in Table \ref{MMR_dataset}, and descriptions are listed below.
\begin{itemize}[leftmargin=0.5cm]
    \item \textbf{FB-IMG-TXT} \cite{MM-TransE} is the KG combined with textual descriptions and images. The triple part is the subset of a classical KG dataset FB15k \cite{FB15k}, and the images are extracted from ImageNet \cite{imagenet}. Compared to it, \textbf{FB15K-237-IMG} \cite{MKGformer} changes the scope of triplets to FB15k-237 \cite{FB15k-237}.

    \item \textbf{IMGpedia} \cite{IMGpedia1} is the KG, which incorporates visual information of the images from the Wikimedia Commons dataset.

    \item \textbf{MKG} \cite{MKG-ALL} consists of two subsets, \textit{i.e.,} \textbf{MKG-Wikipedia} and \textbf{MKG-YAGO}. They both contain visual entities generated by web search engines. But, their triplet parts are extracted from Wikipedia and YAGO, respectively.
    
    \item \textbf{MMKG} \cite{MMKG} offers three subsets, including \textbf{MMKG-FB15k-IMG}, \textbf{MMKG-DB15k}, \textbf{Yago15k-IMG-TXT}, which integrates specific KGs with numeric literals and images.

    \item \textbf{Richpedia} \cite{Rich} is composed of the triplets, textual descriptions, and images. The textual descriptions are derived from Wikidata, and the corresponding visual resources are crawled from the website.
    
    \item \textbf{WN9-IMG-TXT} \cite{IKRL} is the KG combined with textual descriptions and images. The triple part is the subset of a classical KG dataset WN18 \cite{WN18}, and the images are extracted from ImageNet \cite{imagenet}. Compared to it, \textbf{WN18-IMG} \cite{MKGformer} changes the scope of triplets to the whole WN18.
\end{itemize}
\begin{table}[t]
\fontsize{35}{46}\selectfont 
\caption{Typical benchmark datasets for multi-modal knowledge graph reasoning.}
\vspace{-0.2 cm}
\resizebox{\linewidth}{!}{
\begin{tabular}{ccccccc}
\hline
\textbf{Dataset}               & \textbf{Modality} & \textbf{\# Ent.} & \textbf{\# Rel.}      & \textbf{\# Train Facts}  & \textbf{\# Val. Facts} & \textbf{\#   Test. Facts} \\
\hline
\multirow{3}{*}{FB-IMG-TXT \cite{MM-TransE}}    & KG                   & 11,757                & \multirow{3}{*}{1,231}      & \multirow{3}{*}{285,850}     & \multirow{3}{*}{34,863}      & \multirow{3}{*}{29,580}       \\
& TXT                  & 11,757                &                            &                             &                             &                              \\
& IMG                  & 1,175,700              &                            &                             &                             &                              \\
\multirow{2}{*}{FB15k-237-IMG \cite{MKGformer}} & KG                   & 14,541                & \multirow{2}{*}{237}       & \multirow{2}{*}{272,115}     & \multirow{2}{*}{17,535}      & \multirow{2}{*}{20,466}       \\
& IMG                  & 145,410               &                            &                             &                             &                              \\
\multirow{2}{*}{IMGpedia \cite{IMGpedia1}}      & KG                   & 14,765,300             & \multirow{2}{*}{442,959,000} & \multirow{2}{*}{3,119,207,705} & \multirow{2}{*}{-}          & \multirow{2}{*}{-}           \\
& IMG                  & 44,295,900             &                            &                             &                             &                              \\
\multirow{3}{*}{MMKG-FB15k \cite{MMKG}}    & KG                   & 14,951                & \multirow{3}{*}{1,345}      & 592,213                      & -                           & -                            \\
& Numeric      & 29,395                &                            & 29,395                       & -                           & -                            \\
& IMG                  & 13,444                &                            & 13,444                       & -                           & -                            \\
\multirow{3}{*}{MMKG-DB15k \cite{MMKG}}    & KG                   & 14,777                & \multirow{3}{*}{279}       & 99,028                       & -                           & -                            \\
& Numeric       & 46,121                &                            & 46,121                       & -                           & -                            \\
& IMG                  & 12,841                &                            & 12,841                       & -                           & -                            \\
\multirow{3}{*}{MMKG-Yago15k \cite{MMKG}}  & KG                   & 15,283                & \multirow{3}{*}{32}        & 122,886                      & -                           & -                            \\
& Numeric      & 48,405                &                            & 48,405                       & -                           & -                            \\
& IMG                  & 11,194                &                            & 11,194                       & -                           & -                            \\
\multirow{3}{*}{MKG-Wikipedia \cite{MM-RNS}} & KG                   & 15,000                & \multirow{3}{*}{169}       & \multirow{3}{*}{34,196}      & \multirow{3}{*}{4,274}       & \multirow{3}{*}{4,276}        \\
& TXT                  & 14,123                &                            &                             &                             &                              \\
& IMG                  & 14,463                &                            &                             &                             &                              \\
\multirow{3}{*}{MKG-YAGO \cite{MM-RNS}}      & KG                   & 15,000                & \multirow{3}{*}{28}        & \multirow{3}{*}{21,310}      & \multirow{3}{*}{2,663}       & \multirow{3}{*}{2,665}        \\
& TXT                  & 12,305                &                            &                             &                             &                              \\
& IMG                  & 14,244                &                            &                             &                             &                              \\
\multirow{2}{*}{RichPedia \cite{Rich}}     & KG                   & 29,985                & \multirow{2}{*}{3}         & \multirow{2}{*}{119,669,570}  & \multirow{2}{*}{-}          & \multirow{2}{*}{-}           \\
& IMG                  & 2,914,770              &                            &                             &                             &                              \\
\multirow{3}{*}{WN9-IMG-TXT \cite{MM-TransE}}   & KG                   & 6,555                 & \multirow{3}{*}{9}         & \multirow{3}{*}{11,741}      & \multirow{3}{*}{1,319}       & \multirow{3}{*}{1,337}        \\
& TXT                  & 6,555                 &                            &                             &                             &                              \\
& IMG                  & 63,225                &                            &                             &                             &                              \\
\multirow{2}{*}{WN18-IMG \cite{MKGformer}}      & KG                   & 14,541                & \multirow{2}{*}{18}        & \multirow{2}{*}{141,442}     & \multirow{2}{*}{5,000}       & \multirow{2}{*}{5,000}        \\
& IMG                  & 145,410               &                            &                             &                             & \\
\hline
\end{tabular}}
\vspace{-0.15 cm}
\label{MMR_dataset}
\end{table}

\section{Challenge and Opportunity}
According to previous analyses of the existing KGR models, we point out several promising directions for future works.

\subsection{Out-of-distribution Reasoning}
In the real-world scenario, new entities and relations are continuously emerging in the KGs, which are under-explored in the original KGs. Reasoning on the facts with these under-explored elements is called out-of-distribution reasoning, which raises higher requirements for the KGR model design. 
Some recent attempts provide potential solutions for inferring unseen entities, which are known as inductive reasoning models, such as \cite{GraIL,TACT,CoMPILE,SNRI}. These models mine the logic rules underlying the graph structure without considering the specific meaning of entities, which achieve promising performances. As for the unseen relation inference, few-shot KGR models \cite{CSR,MetaTKGR,Meta-iKG} tend to improve the generalization ability of models so that the trained model can scale well to the unseen relations with a small amount of facts. 
In other words, these few-shot KGR models can quickly learn new tasks according to the previously learned similar knowledge. Besides, BERTRL \cite{BERTRL} tries to handle this case based on their textual semantics calculated by language models. While the performance of these models would drop drastically when language models are not finely trained. In conclusion, the KGR models for out-of-distribution reasoning tasks are still in an early stage, which is worth exploring in-depth in the future.

\subsection{Large-scale Reasoning}
The industrial KGs are generally large-scale, which requires more efficient KGR models. To this end, some existing works try to optimize the propagation procedures in a progressive manner \cite{astar}. For instance, NBF-net \cite{NBF-net} integrates the bellman-ford algorithm to substitute the original DFS-based aggregation procedure in GNN-based KGR models. Moreover, A$^*$Star \cite{astar} Net further optimizes the aggregation procedure with the greedy algorithm. Besides, the idea of graph clustering \cite{deep_graph_clustering_survey,yaming_1,xihong_1} is also used for it. For example, CURL \cite{CURL} first separates the KGs into different clusters according to the entity semantics and then fine-grains the path-finding procedure into two-level, \textit{i.e.,} the intra-cluster level and the inter-cluster level. It reduces the unnecessary searching for the whole graphs. Similarly, many works perform reasoning on sub-graphs instead of complete graphs, such as GraIL \cite{GraIL}, CSR \cite{CSR} etc. But most of them sacrifice the precision of inference, which may still be explored for more all-around models.

\subsection{Multi-relational Reasoning}
\begin{figure}[t]
\centering
\includegraphics[width=0.49\textwidth]{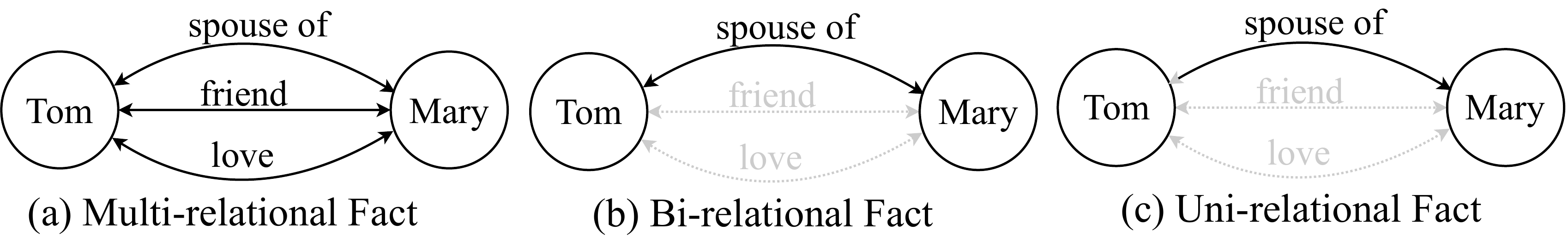}
\caption{Comparison of multi-relational, bi-relational and uni-relational facts.}
\label{Multi-uni}  
\vspace{-10pt}
\end{figure}

The situation that multi-relational facts exist between two entities is common in KGs as shown in Figure \ref{Multi-uni} (a). However, they are more diverse in structure and more complex in semantics compared to uni-relational and bi-relational facts as shown in Figure \ref{Multi-uni} (b) and (c). 
Thus, the existing KGR models mainly focus on uni-relational and bi-relational facts and even usually treat multi-relational facts as uni-relational and bi-relational facts by omitting some of the facts. The KGR models in such a manner cannot accurately model real situations and lose lots of meaningful semantic information, leading to insufficient expressive ability. In the future, it is necessary to study how to leverage multi-relational facts to enhance reasoning ability.

\subsection{Multi-modal Reasoning}
Knowledge reasoning based on the fusion of multi-source information can reduce the disconnectedness and sparsity of knowledge graphs by combining a text corpus or additional information in other modalities. Knowledge reasoning based on the fusion of data in multiple modalities can complement each other’s advantages and improve reasoning performance. However, existing multi-modal KGR models are still at an early stage. They still tend to directly concat the embeddings in different modalities together for final score calculation. Such simple fusion modes have shown their promising performances while developing more fine-grained and scalable modes is still worthwhile. For instance, an adaptive fusion mode, which weighs the importance of different modalities, is worthwhile exploring.

\subsection{Explainable Reasoning}
Explainability is a common and important issue for deep learning models in various fields. Although KGR models generally are more explainable, it is still worthwhile exploring more in this topic, especially for embedding-based KGR models. Nowadays, more and more KGR models are developed based on neural networks, such as GNN \cite{graphPAMI}. Most of them have the great expressive ability but suffer from explainability. Compared to them, rule-based and path-based KGR models are more explainable but computation-consuming and less expressive \cite{TransferPAMI}. To achieve a good trade-off between expressive ability and explainability, there exist some attempts to integrate the embedding-based models with rule-based and path-based models, such as ARGCN \cite{MBE}. It builds the reward function based on the embeddings generated by the RGCN \cite{RGCN}, which makes those path-based models more explainable. However, most of these attempts are still rough.

\subsection{Knowledge Graph Reasoning Application}
Although a large amount of KGR methods have been proposed in recent years, demonstrating the great potential of KGR in theoretical fields, the applications of KGR still need to be studied more \cite{KFPAMI}. Nowadays, knowledge graphs are commonly used in many downstream applications, such as medicine, finance, plagiarism detection, etc. Medical knowledge reasoning models aim to assist doctors in diagnosing diseases from electronic medical records. For example, \cite{medical1} and \cite{medical2} both perform reasoning on the KG constructed from the electronic medical database. The pre-trained language models, such as Bert, are leveraged to generate textual embedding of entities, which is proven effective in existing multi-modal KGR models. Besides, KGR models can also help with Anti-fraud detection, which is an important task in the finance field. For instance, \cite{finacial} proposes a case-based reasoning method to assist people in verifying the information to discriminate against fraud in advance. Additionally, \cite{chaoxi} executes plagiarism detection by conducting the KGR approach in a continuous learning manner.

\subsection{Knowledge Graph and Large Language Model}
Large language models (LLMs) \cite{liu2023summary}, \textit{i.e.,} ChatGPT, GPT-4, are very popular this year, which have huge impacts due to their promising reasoning capacity and generalizability \cite{pan2023unifying}. However, these models still suffer from two issues, \textit{i.e.,} (1) the poor explainability, and (2) poor scalability when handling new data, which may be solved when if well cooperated with KGR models. For example, QA-GNN \cite{yasunaga2021qa} first attempts to use LLMs for text preprocessing and further guides the reasoning step on the KGs. Besides, DRAGON \cite{DRAGON} is an LLM-guided logical reasoning method for multi-modal KGR. Meanwhile, since more and more data are being trained, some researchers suggest that LLM is a more general KG in the future, which can also achieve functions, like indexing, reasoning, storing, etc. Various hypotheses about the connections between LLM and KG are all reasonable, and we cannot say which one is more valuable. However, we can be sure that exploring and researching between LLM and KG will also be one of the hot spots in the future.

\section{Conclusion}
Our survey thoroughly reviews the existing KGR models based on a bi-level taxonomy, \textit{i.e.,} top level (graph types), and base level (techniques, scenarios). Three graph types (\textit{i.e.,} static, temporal, multi-modal KGs), fourteen techniques, and four reasoning scenarios are included, which provides systematical reviews for KGR. Besides, we summarize the challenges of knowledge graph reasoning and point out some potential opportunities which will enlighten the readers. The corresponding open-source repository for the collection of {180} state-of-the-art KGR models (\textit{i.e.,} papers, and codes) and {67} typical datasets are shared on GitHub to convenient the community.

\ifCLASSOPTIONcaptionsoff
  \newpage
\fi

\bibliographystyle{IEEEtran}
\bibliography{TKDE}

\end{document}